\documentclass[a4paper,fleqn]{cas-dc}

\usepackage[numbers]{natbib}

\usepackage{lineno}

\usepackage{algorithmic}
\usepackage{algorithm}
\usepackage{array}
\usepackage[caption=false,font=normalsize,labelfont=sf,textfont=sf]{subfig}
\usepackage{textcomp}
\usepackage{stfloats}
\usepackage{url}
\usepackage{verbatim}
\usepackage{graphicx}
\usepackage{cite}
\usepackage{booktabs}

\usepackage{amsmath,amssymb,amsfonts}%
\usepackage{amsthm}%
\usepackage{mathrsfs}%
\usepackage{multirow}

\def\tsc#1{\csdef{#1}{\textsc{\lowercase{#1}}\xspace}}
\tsc{WGM}
\tsc{QE}

\begin{document}
\let\WriteBookmarks\relax
\def\floatpagepagefraction{1}
\def\textpagefraction{.001}

\shorttitle{}    

\shortauthors{}  


\title [mode = title]{A Comprehensive Survey on Synthetic Infrared Image synthesis}



%

\author[1]{Avinash Upadhyay}
\fnmark[1]
\ead{avinre@gmail.com}
\author[2]{Manoj Sharma}
\ead{mksnith@gmail.com}
\cormark[2]
\author[3]{Prerana Mukherjee}
\author[4]{Amit Singhal}
\author[5]{Brejesh Lall}

\affiliation[1]{organization={Electronic Science Department, SEAS, Bennett University},
            city={Gautam Buddha Nagar},
            postcode={201310}, 
            state={Uttar Pradesh},
            country={India}}
\affiliation[2]{organization={Electronic Science Department, SEAS, Bennett University},
            city={Gautam Buddha Nagar},
            postcode={201310}, 
            state={Uttar Pradesh},
            country={India}}

\affiliation[3]{organization={School of Engineering, Jawaharlal Nehru University},
            city={New Delhi},
            postcode={110067}, 
            state={Delhi},
            country={India}}
\affiliation[4]{organization={Department of Electronic Science, Netaji Subhash University of Technology},
            city={New Delhi},
            postcode={110078}, 
            state={Delhi},
            country={India}}
\affiliation[5]{organization={Electrical Engineering Department, IIT Delhi},
            city={New Delhi},
            postcode={110016}, 
            state={Delhi},
            country={India}}
















\nonumnote{}

\begin{abstract}
Synthetic infrared (IR) scene and target generation is an important computer vision problem as it allows the generation of realistic IR images and targets for training and testing of various applications, such as remote sensing, surveillance, and target recognition. It also helps reduce the cost and risk associated with collecting real-world IR data. This survey paper aims to provide a comprehensive overview of the conventional mathematical modelling-based methods and deep learning-based methods used for generating synthetic IR scenes and targets. The paper discusses the importance of synthetic IR scene and target generation and briefly covers the mathematics of blackbody and grey body radiations, as well as IR image-capturing methods. The potential use cases of synthetic IR scenes and target generation are also described, highlighting the significance of these techniques in various fields. Additionally, the paper explores possible new ways of developing new techniques to enhance the efficiency and effectiveness of synthetic IR scenes and target generation while highlighting the need for further research to advance this field.
\end{abstract}


\begin{highlights}
\item The paper surveys recent research in infrared scene simulation, contrasting traditional physics-based computational methods with innovative deep learning approaches that generate infrared imagery from RGB inputs.
\item A comparison of leading infrared scene simulation tools, is presented, highlighting their capabilities, applications, and advancements in simulation techniques.
\item The paper provides an in-depth examination of the fundamental principles of infrared physics, including the characterization of infrared sources, atmospheric effects on infrared transmission, and the detector.
\end{highlights}



\maketitle

\section{Introduction}
\label{sec:intro}
Infrared (IR) imaging is a transformative technology with profound implications across a multitude of fields. It penetrates the veil of darkness, sees through smoky and hazy conditions, and unveils a world invisible to the naked eye. Ranging from defence and surveillance systems using IR imaging for target tracking and night vision, to healthcare leveraging it for non-invasive diagnostics, and meteorology employing it for weather prediction and climate study - the applications of IR imaging are myriad and varied. The power of IR imaging lies in its ability to provide unique information about our environment that is unattainable through other imaging modalities.

Despite its enormous potential, the development and deployment of IR systems face a significant bottleneck - the scarcity of suitable IR data. Training robust and reliable IR systems requires large, diverse, and high-quality datasets. However, obtaining such datasets in real world is challenging due to privacy concerns, logistical issues, and the difficulty in capturing a wide range of scenarios and conditions. This data limitation impedes the progression of IR technology, restricting its ability to learn, adapt, and generalize to novel environments.

Addressing this challenge is where the potential of synthetic Infrared imagery and video synthesis shines. Through advanced computational models and simulations, synthetic IR imaging can generate large volumes of realistic, diverse, and controlled IR data. This technology can emulate a multitude of real-world scenarios, from varying atmospheric conditions to different material properties, thereby providing a rich, flexible, and ethical source of data for training IR systems. It paves the way for IR systems to learn and adapt effectively, bridging the gap between the scarcity of real-world IR data and the data-hungry nature of modern IR systems.
This comprehensive survey has a dual purpose: to aggregate both traditional and learning-based methods for synthetic Infrared (IR) generation across the full spectrum of the Infrared Band - from Near-Infrared (NIR), through Mid-Infrared (MIR), to Far-Infrared (FIR), spanning a wavelength range of 0.7 to 14 micrometers. It is an endeavor to consolidate the wealth of research conducted in the realm of synthetic IR generation for various applications. To the best of our knowledge, this survey is the first of its kind, offering a holistic compilation of methods available for synthetic Infrared image and video generation. We envisage that this survey will be invaluable for individuals and organizations involved in this field, enabling them to identify and understand the array of algorithms currently in use. We believe that this in-depth overview will stimulate new research, encouraging researchers to effectively address existing challenges and pioneer innovative solutions in this domain. Additionally, this survey will provide industry professionals with a spectrum of potential methods for synthetic IR generation, assisting them in identifying the most suitable algorithm for their specific use cases. Through this survey, we aim to foster advancement and innovation in the field of synthetic Infrared imaging, bridging the gap between academic research and industry application.

This survey paper delves into the realm of synthetic Infrared image and video synthesis, covering its principles, applications, methodologies, and challenges. We initiate our exploration with the `Basics of Infrared Imagery', shedding light on the fundamental physics and mathematics underpinning this technology in sec. \ref{sec:basics}. The paper then navigates through the diverse `Applications of Synthetic Infrared Images' in sec. \ref{sec:tools} and provides insights into the available `Datasets' in sec. \ref{sec:dataset}. We subsequently elucidate the `Methodologies' for generating synthetic IR images and videos in sec. \ref{sec:method}, and address the associated `Challenges' in sec. \ref{sec:challenges}. In the final segment, we tie together our discussions in the `Conclusion', reflecting on the current state and future trajectory of synthetic Infrared image/video synthesis in sec. \ref{sec:conclu}. As we traverse this landscape, we aim to provide a comprehensive resource for researchers, practitioners, and enthusiasts in this field.

\section{Basics of Infrared Imagery}
\label{sec:basics}
This section provides a brief overview and mathematical preliminaries of infrared (IR) imagery. It discusses blackbody and graybody radiation, methods of detecting IR radiation, the different types of IR, and tools and algorithms for modeling the atmospheric transfer function.

\begin{table*}[h]
\centering
\begin{tabular}{|c|c|p{4cm}|}
\hline
\textbf{Type of IR} & \textbf{Wavelength Band} & \textbf{Usage}\\
\hline
Near-IR (NIR) & 0.7 - 0.9 µm & Telecommunications, fiber optics\\
Short-Wavelength IR (SWIR) & 0.9 - 2.5 µm & Night vision, remote sensing\\
Mid-Wavelength IR (MWIR) & 3 - 5 µm & Thermal imaging, missile guidance\\
Long-Wavelength IR (LWIR) & 8 - 14 µm & Thermal imaging, night vision, surveillance\\
Far-IR (FIR) & 15 - 1000 µm & Astronomy, environmental monitoring, thermal efficiency analysis\\
\hline
\end{tabular}
\caption{Types of IR, their Wavelength Bands, and Usage}
\label{tab:IR_types}
\end{table*}

\subsection{Types of IR}

Infrared (IR) radiation is a diverse range of electromagnetic waves, each with unique characteristics and applications. The near-IR (NIR) spectrum, spanning 0.7-0.9 µm, is widely used in telecommunications and fiber optics, enabling fast and reliable data transmission over long distances. Short-wavelength IR (SWIR), operating within 0.9-2.5 µm range, is employed in night vision devices, allowing users to navigate in low-light conditions, and in remote sensing, where it helps gather data on the Earth's surface. Mid-wavelength IR (MWIR) and long-wavelength IR (LWIR) are both utilized in thermal imaging, with MWIR (3-5 µm) also used in missile guidance systems due to its ability to penetrate atmospheric obstacles. LWIR (8-14 µm), on the other hand, is used in surveillance and night vision applications, where its longer wavelengths can detect temperature differences in objects. Finally, far-IR (FIR), covering 15-1000 µm, has applications in astronomy, where it helps study the formation of stars and galaxies, as well as in environmental monitoring and thermal efficiency analysis, where it can detect subtle changes in temperature and energy patterns. Different types of IR spectrum is tabulated in table \ref{tab:IR_types} along with the popular use cases of those spectrums.

\subsection{Infrared Radiometry and Detection}\label{sec2}

\begin{figure*}
    \centering
    \includegraphics[width=1\linewidth]{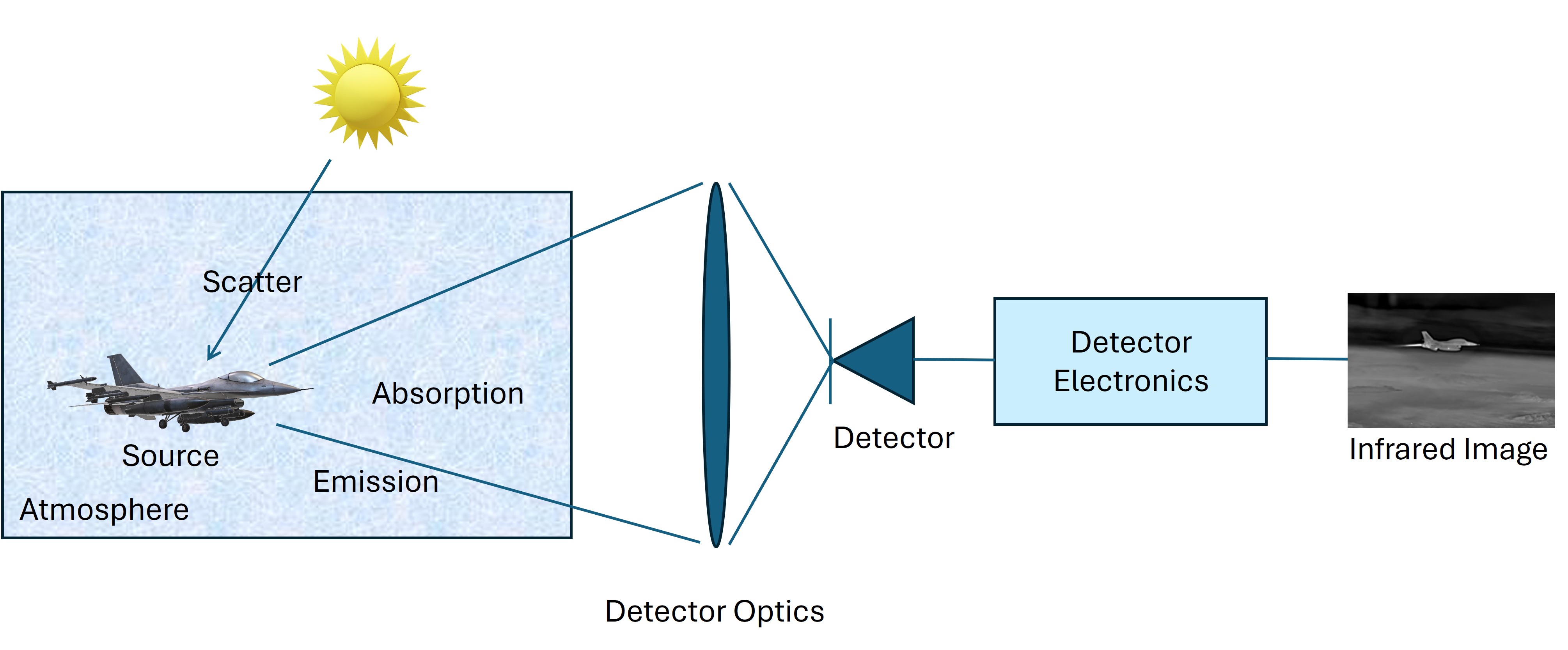}
    \caption{Pictorial representation of an Infrared System.}
    \label{fig:IR_SYSTEM}
\end{figure*}

The operation of infrared imaging systems is predicated on the detection of thermal radiation emitted by objects, a phenomenon governed by the principles of radiometry. To accurately interpret the infrared radiation received by a sensor, a comprehensive understanding of the underlying radiometric physics is essential. This includes the calculation of surface temperature through the solution of heat equilibrium equations, the determination of spectral radiance based on the surface temperature, and the subsequent calculation of radiance intensity. Furthermore, the atmosphere plays a significant role in attenuating the radiation, necessitating the modeling of atmospheric transmittance to accurately predict the radiation received by the sensor. This subsection provides an in-depth examination of the radiometry and detection aspects of infrared imaging systems, encompassing the calculation of surface temperature, spectral radiance, and radiance intensity, as well as the modeling of atmospheric transmittance and the conversion of radiance intensity to voltage. A thorough grasp of these fundamental concepts is crucial for the design and optimization of infrared imaging systems for a diverse range of applications, including thermal imaging, surveillance, and environmental monitoring. A diagram representing the Infrared system is presented in the figure \ref{fig:IR_SYSTEM}.

\subsubsection{Heat Equilibrium equation and surface temperature calculations}

The ability of a body to radiate is closely related to its ability to absorb radiation since the body is in thermal equilibrium with its surroundings at a constant temperature; it must absorb and radiate energy at the same rate.  The thermal equilibrium equation for an object can be described as the summation of absorbed incident radiations from the sun and environment, conduction of internal heat source is equal to the radiance emitted by the blackbody and the heat convection and other effects \citep{Choi2009,Yu1998,willers2011}. 

\begin{equation}
    Q_d = Q_i + Q_{sun} + Q_{env} + Q_{conv}
\end{equation}

Where \(Q_d\) is heat conduction into the object, \(Q_i\) is the internal heat source, \(Q_{sun}\) is the absorbed solar energy which constitutes of both direct and diffuse components, \(Q_{env}\) is the energy absorbed from the environmental factors. The heat conduction component within the object can be written in terms of specific heat capacity \(C\),  linear density of the object \(\rho\), and depth of heat penetration \(h\). Specific heat capacity and density of the object are material properties.

\begin{equation}
\label{heat_equilibrium_temperature}
    C\rho h \frac{dT}{dt} = Q_i + Q_{sun} + Q_{env} + Q_{conv}
\end{equation}

Solving the above equation gives the Temperature of the surface \(T\) of the object. 

\subsubsection{Spectral radiance of body}

The spectral radiance of the black body for specific wavelength $\lambda$ and temperature $T$ is given as follows, 

\begin{equation}
S_r = \frac{c_1}{\lambda^5}\frac{1}{\exp\{{\frac{c_2}{\lambda T}}\}-1} d\lambda \quad W/cm^2\mu sr
\end{equation}

Where $c_1=1.191\times104[\frac{W\mu m^4}{cm^2 sr}]$, and $c_2=1.428\times104[\mu mK]$ are radiance constants. The radiance intensity $L(T)$ of an IR wavelength bandwidth for a blackbody will be given as,

\begin{equation}
L(T) = \int_{\lambda_2}^{\lambda_1} \frac{c_1}{\lambda^5}\frac{1}{\exp\{{\frac{c_2}{\lambda T}}\}-1} d\lambda \quad W\mu sr/cm^2
\end{equation}

And for a graybody whose spectral emissivity is less than 1, the equation will be, 

\begin{equation}
L(T) = \int_{\lambda_2}^{\lambda_1} \frac{\epsilon(\lambda)c_1}{\lambda^5}\frac{1}{\exp\{{\frac{c_2}{\lambda T}}\}-1} d\lambda \quad W\mu sr/cm^2
\end{equation}

Where $\epsilon(\lambda)$ is the spectral emissivity of the object.

\subsubsection{Radiance to Detector Voltage}

If such radiation falls on the IR detector with an area of $A_p$ and the solid angle of the detector at the target is $\Omega$, the output voltage $V_{det}$ generated by the detector will be, 

\begin{equation}
    V_{det} = A_p\Omega L(T) \int_{\lambda_2}^{\lambda_1} \tau_{amb}(\lambda)\tau_{opt}(\lambda)\mathfrak{R}(\lambda)d\lambda \quad V
\end{equation}
 
Where $\tau_{amb}(\lambda)$ represents the atmospheric transfer function and $\tau_{opt}(\lambda)$ represents the transmittance of detector optics. $\mathfrak{R}$ is the spectral responsivity of the detector with the unit $[V/W]$. The responsivity function, denoted as $R(\lambda, x, y, t)$, describes the system's response to an input signal, where $\lambda$ represents the spectral distribution, $x$ and $y$ represents the spatial distribution, and $t$ represents the temporal distribution. In an ideal system, the responsivity function would be a linear and shift-invariant function of the input signal, allowing for straightforward prediction of the system's response. However, in practice, the responsivity function is often nonlinear and shift-variant, due to factors such as detector nonlinearity, optical aberrations, and electronic noise. If we assume these variables as shift-invariant system, integral values of the wavelength band will have constant value. Since, the detector area and the solid angle is also constant we can safely say, 

\begin{equation}
    V_{det} \propto L(T)
\end{equation}

The spectral radiance intensity is the function of the spectral emissivity $\epsilon(\lambda)$ and surface temperature. The emissivity is constant for the surface. Hence, it can be derived that the Voltage generated by the detector is directly proportional to the temperature of the surface. 

The equation relating radiance to pixel brightness in IR imaging depends on several factors, including the specific MWIR imaging system used, the target’s characteristics, and the atmospheric conditions at the imaging time.  

In general, the process of converting radiance (measured in watts per square meter per steradian) to pixel brightness (measured in digital counts or units) involves several steps, including: 
    \begin{enumerate}[1.]
        \item \textbf{Calibration:} The IR imaging system must be calibrated using a known radiance source to establish a relationship between radiance and pixel brightness. 
    
        \item \textbf{Correction for atmospheric effects:} The radiance measured by the imaging system may be affected by scattering, absorption, and other atmospheric effects that can reduce image quality. These effects can be corrected using atmospheric correction models and algorithms. 
    
        \item \textbf{Temperature conversion:} The radiance can be converted to temperature using a radiometric conversion equation that considers factors such as the target's emissivity and the imaging system's spectral response. 
    
        \item \textbf{Display:} Finally, the temperature values can be mapped to pixel brightness values using a lookup table or color scale to represent the IR image visually. 
    \end{enumerate}
    
Each step's specific equations and algorithms will vary depending on the IR imaging system and application. 
IR wavelength band conversion is an extremely difficult and challenging area of research because materials have different emissivity, reflectivity and transmissivity that varies with the IR wavelength band. The variance is strongly affected by water vapour and carbon dioxide amongst atmospheric gases because these components have a specific absorption/emission spectrum according to the wavelength band.

\section{Tools and Algorithms for modelling atmospheric transfer function}
\label{sec:tools}
Atmospheric transmission or atmospheric transfer function refers to the process by which electromagnetic waves, including light and other forms of radiation, pass through the Earth's atmosphere. The atmosphere is made up of a mix of different gases, particles, and water vapour, all of which can interact with electromagnetic waves in various ways. These interactions can include absorption, where the energy of the wave is taken up by the atmospheric components, and scattering, where the direction of the wave is changed.
In the context of IR imaging, atmospheric transmission is particularly relevant because the atmosphere can absorb specific wavelengths of IR radiation more than others. This is due to the presence of gases like water vapour and carbon dioxide, which can absorb specific frequencies of IR radiation. This effect is known as atmospheric absorption or atmospheric attenuation.
Accurate measurement of the atmospheric transfer function is crucial in synthetic IR imaging for realism and precision. This accurate modelling enhances the quality of synthetic images, creating a more representative training dataset for machine learning models. Furthermore, it improves the generalizability of these models to real-world conditions and allows for effective performance evaluation of IR imaging systems, ensuring they can effectively interpret real-world IR data.

\subsubsection{DISORT (Discrete Ordinate Radiative Transfer)}
This method is a widely used radiative transfer model that can simulate the transmission of radiation through a horizontally stratified atmosphere. DISORT uses a discrete ordinate method to solve the radiative transfer equation and can account for scattering and absorption by clouds and aerosols. 
\subsubsection{RTTOV (Radiative Transfer for TOVS)}
 This method was developed by the European Centre for Medium-Range Weather Forecasts (ECMWF) and is used to simulate the transmission of MWIR radiation through the Earth's atmosphere for weather forecasting applications. RTTOV uses a fast radiative transfer algorithm to calculate the radiative transfer coefficients based on atmospheric profiles.
 \subsubsection{ARTS (Atmospheric Radiative Transfer Simulator)}
 This method is a flexible radiative transfer model that can simulate the transmission of MWIR radiation through complex atmospheric conditions, such as clouds and aerosols, and can be used to simulate the performance of remote sensing instruments. 
\subsubsection{TES (Transmittance Estimation from the Surface)}
 This method is a simplified model that estimates the transmittance of MWIR radiation through the Earth's atmosphere based on the surface temperature and atmospheric water vapour content. TES is often used for quick estimates of radiation transmittance for remote sensing applications. 
\subsubsection{LOWTRAN and MODTRAN}
LOWTRAN (Low-Resolution Transmission) and MODTRAN (MODerate resolution atmospheric TRANsmission) are widely used methods for simulating the transmission of MWIR radiation through the Earth's atmosphere. LOWTRAN was developed by the US Air Force and is a simplified model that calculates radiation transmission based on the atmospheric profile and the concentration of atmospheric constituents, such as water vapour, carbon dioxide, and ozone. LOWTRAN assumes a horizontally homogeneous atmosphere and provides low spectral resolution. MODTRAN, on the other hand, is a more sophisticated model developed by the US Air Force and Spectral Sciences, Inc. It provides moderate spectral resolution and can account for more complex atmospheric conditions, such as horizontal and vertical variations in atmospheric constituents and the effects of clouds and aerosols on radiation transmission. MODTRAN can also simulate the reflection and emission of radiation from the Earth's surface and can be used to predict the performance of remote sensing instruments, such as satellites or airborne sensors. Both LOWTRAN and MODTRAN are widely used by researchers and practitioners in atmospheric science, remote sensing, and defense applications to simulate the transmission of MWIR radiation through the Earth's atmosphere and to evaluate the performance of MWIR remote sensing instruments.

\subsubsection{ATRAN\citep{ATRAN}}
ATRAN (Atmospheric TRANsmission) is a software module developed for accurately modeling the transmission of electromagnetic radiation through Earth's atmosphere. It considers factors such as altitude, temperature, pressure, and gas concentrations to compute synthetic spectra of atmospheric transmission. Used in fields like meteorology, remote sensing, and telecommunications, ATRAN aids in interpreting sensor data, designing communication systems, and generating realistic synthetic IR images.

\begin{figure}[h]
\centering
\includegraphics[width=0.5\textwidth]{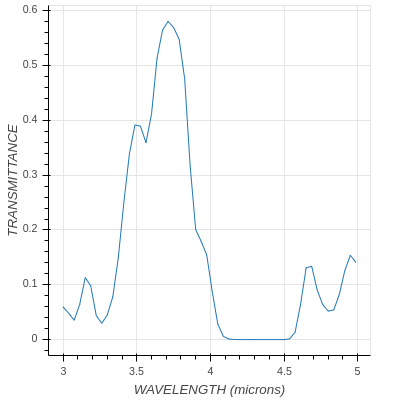}
\caption{Atmospheric transmission pattern generated by MODTRAN module for the wavelength range of MWIR 3$\mu m$ to 5$\mu m$. Atmosphere Model used is US Standard 1976, ground temperature is 294.2K, Aerosol model used is Urban with a visibility of 23km, sensor altitute is 5km and zenith is 90 degrees.}
\label{atran_img}
\end{figure}

\section{Datasets}
\label{sec:dataset}
In this section, we explore a diverse range of infrared datasets, which can be used to generate the synthetic IR scenes and targets. The datasets on the Infrared can be categorized into three types: i) standalone Infrared datasets, ii) RGB-Infrared unpaired datasets, and iii) RGB-Infrared paired datasets. Standalone Infrared Datasets consists exclusively of infrared images captured using IR sensors. They are particularly useful for applications focusing solely on infrared imaging or for developing algorithms that leverage the unique characteristics of IR data. RGB-Infrared Unpaired Datasets comprise both RGB and infrared images; the images in these datasets are not paired, which means that corresponding RGB and infrared images may not share the same scene or capture time. These datasets are valuable for exploring the strengths of RGB and infrared imaging and for developing algorithms that can process and analyze both data types independently. In the RGB-Infrared Paired Datasets, RGB and infrared images are strictly paired, meaning they share the same scene and capture time. This pairing allows researchers to develop and evaluate algorithms that fuse RGB and infrared data, capitalizing on the complementary information provided by both imaging modalities.

\begin{figure}
    \centering
    \includegraphics[width=0.7\linewidth]{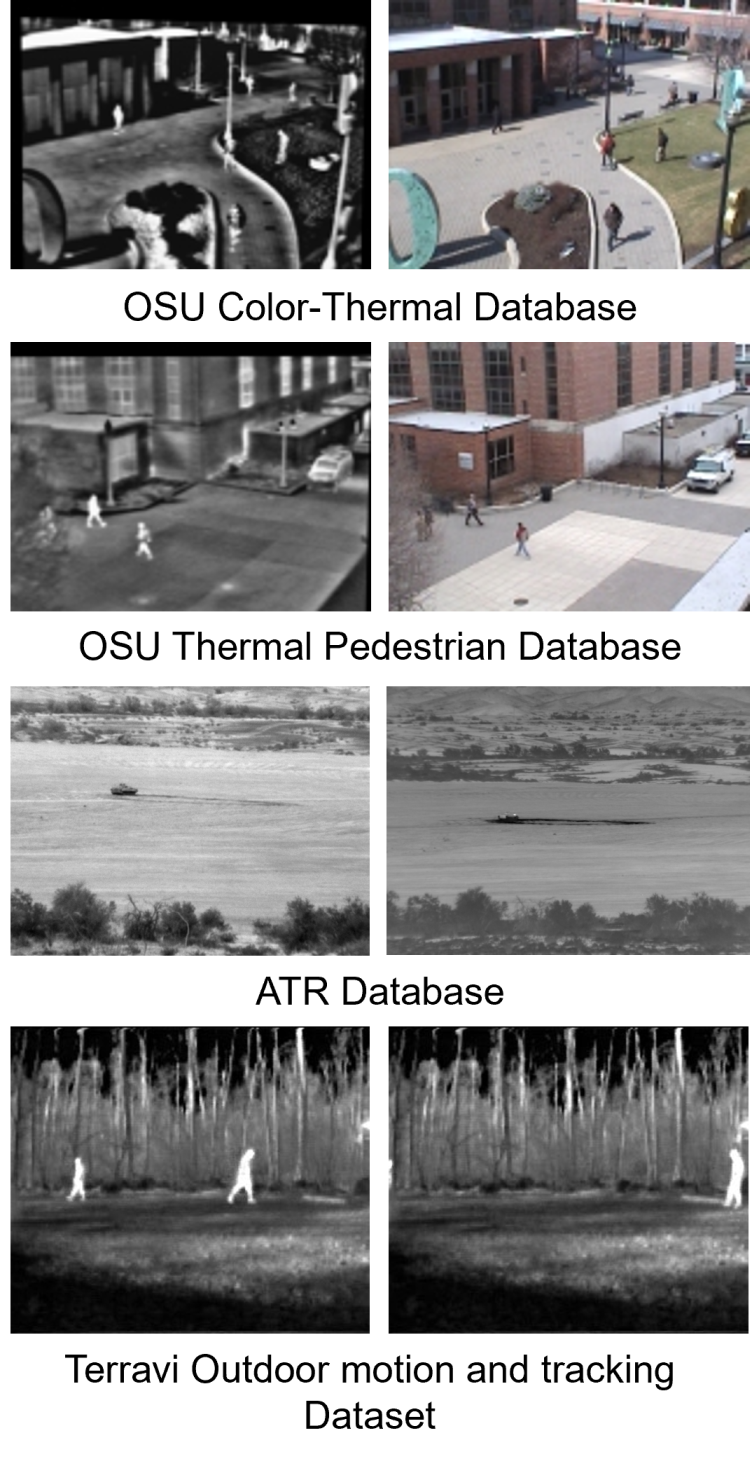}
    \caption{Image snippets from different datasets.}
    \label{fig:dataset}
\end{figure}

Further, the datasets discussed in this section cover a broad spectrum of imaging scenarios, encompassing aerial, outdoor-non-aerial, and indoor images. Aerial infrared datasets consist of images captured from elevated platforms such as satellites, aeroplanes, or unmanned aerial vehicles (UAVs) like drones. These datasets provide valuable information for various applications, including environmental monitoring, disaster management, agriculture, urban planning, and land use analysis. Aerial infrared images reveal insights not readily discernible in visible light images, such as temperature variations, moisture content, and vegetation health. The outdoor-Non-Aerial category includes infrared datasets captured at ground level in various outdoor environments. These datasets can cover diverse scenarios, such as pedestrian and vehicle detection, road and traffic monitoring, infrastructure assessment, and wildlife observation. Outdoor-non-aerial infrared images can be particularly beneficial in low-light or obscured conditions, where traditional RGB imaging may struggle to provide sufficient information for computer vision tasks. Indoor infrared datasets comprise images taken within enclosed spaces, such as buildings, homes, factories, or other structures. These datasets can be used for applications like thermal anomaly detection, energy efficiency analysis, surveillance and security, and human activity recognition. Indoor infrared imaging can reveal hidden details, such as heat signatures, moisture levels, or insulation issues, which can be critical for assessing building performance or detecting potential safety hazards. Following are the datasets available in the public domain:

\subsection{OSU Databases}
The datasets under discussion were meticulously prepared and shared by researchers at Ohio State University.
\subsubsection{OSU Thermal Pedestrian Database \citep{OSU_THERMAL_PEDESTRIAN}}
Comprised of thermal images in the wavelength range of 7-14 $\mu m$, this standalone thermal dataset was proposed for developing person detection algorithms. The images were captured using a Raytheon 300D thermal sensor, equipped with a high-quality 75mm lens, ensuring accurate and detailed infrared data. The dataset contains a total of 10 video sequences, which are further divided into 284 individual images. Each image is presented in 8-bit grayscale bitmap format, with a resolution of 360$\times$240 pixels.  Specifically designed for person detection applications, this dataset enables the development and evaluation of algorithms that can accurately identify and track individuals in a variety of environments and conditions.

\subsubsection{OSU Color-Thermal Database\citep{OSU_COLOR_THERMAL}}
This dataset focuses on the fusion of color and thermal imagery, with specific applications in fusion-based object detection using both color and thermal data. The dataset was collected using two sensors: a Raytheon PalmIR 250D thermal sensor with a 25mm lens and a Sony TRV87 Handycam color sensor. These cameras were mounted adjacent to each other on a tripod at two separate locations, approximately three stories above the ground, with manual control over gain and focus settings. The dataset consists of six color and thermal sequences, with three captured at each location. In total, there are 17,089 images in the dataset. The thermal images are in 8-bit grayscale bitmap format, while the color images are in 24-bit color bitmap format. Each image has a resolution of 320x240 pixels, and the sampling rate is approximately 30Hz. The color and thermal images were registered using homography, a technique that relies on manually-selected points to align the images accurately. This registration process ensures that the corresponding color and thermal images share the same scene and capture time, making it possible to develop and evaluate fusion-based object detection algorithms. The dataset contains the thermal images in the wavelength range of 7-14 $\mu m$.

\subsection{ATR Algorithm Development Image Database \citep{ATR}}
The US Army NVESD has created the ATR (Automatic Target Recognition) Database for ATR algorithm developers. It contains ~600000 visible and MWIR 3-5 $\mu m$ images of various targets and backgrounds, ground truth data, meteorological data and related information. The image size of MWIR The database covers different target types at different distances and angles, such as people, military vehicles, and civilian vehicles. The MWIR and RGB images both are captured in 16 bit format and has resolution of 640x480. The database also provides target temperature differentials, meteorological data, target photographs, and a user’s guide. The images were captured using commercial cameras in the MWIR and visible bands. It is one of the largest and only available public data in the MWIR band. This dataset is exclusive and available only to researchers in NATO member countries.

\subsection{Terravi Databases \citep{TERRAVIC_IR_DATA}}
This dataset focuses on detection and tracking using thermal imagery. The images were captured using a Raytheon L-3 Thermal-Eye 2000AS sensor with wavelength sensitivity of 7-14 $\mu m$, which specializes in acquiring detailed thermal data. The dataset comprises 18 thermal sequences in total, covering a diverse range of scenarios to accommodate various applications and challenges. These scenarios include 11 outdoor motion and tracking scenes, 1 outdoor house surveillance scene, 1 indoor hallway motion scene, 1 plane motion and tracking scene, 2 underwater and near-surface motion scenes, and 2 uneventful background motion scenes. This wide variety of scenarios ensures that the dataset is suitable for exploring different detection and tracking tasks using thermal imagery. The images in the dataset are in 8-bit grayscale JPEG format. Each image has a resolution of 320x240 pixels, providing sufficient detail for computer vision tasks. By offering a comprehensive collection of thermal images across various scenarios, this dataset enables the development and evaluation of advanced detection and tracking algorithms that leverage the unique advantages of thermal imaging.

\subsection{CSIR-CSIO Moving Object Thermal Infrared Imagery Dataset (MOTIID)
\citep{CSIO0,CSIO1,CSIO2}}
This dataset focuses on moving object detection in thermal infrared imagery, targeting various subjects such as pedestrians, vehicles, and animals. The images were captured using a thermal infrared camera mounted on a tripod at a height of approximately 4 feet. Additional sensor details can be found in the referenced materials. The dataset contains 18 thermal sequences, encompassing a diverse range of moving targets, including two different models of 4-wheelers (Ambassador and Innova), a 3-wheeler (auto-rickshaw), a 2-wheeler (motorcycle), humans walking at varying distances, a strolling dog, and a flying bird. These diverse scenarios enable researchers and developers to explore and develop robust detection algorithms for various moving objects in thermal infrared imagery. The images have a resolution of 640x480 pixels, and the sampling rate is 10Hz. The duration of each thermal video sequence varies between 4-22 seconds, with each sequence featuring one or more moving targets entering and exiting the camera's field of view. This comprehensive dataset provides a valuable resource for the development and evaluation of advanced moving object detection algorithms using thermal infrared imaging.

\subsection{Maritime Imagery in the Visible and Infrared Spectrums \citep{Zhang2015VAISAD}}
The VAIS dataset is a collection of simultaneously acquired unregistered thermal and visible images of ships acquired from piers. It is suitable for object classification research. The dataset includes 2865 images, of which 1242 are infrared images and 1623 are visible images. There are 1088 pairs of images in the dataset, each of which contains an infrared image and a visible image of the same ship. The dataset includes 264 unique ships, of which 154 are night IR images. The dataset is divided into 6 basic categories: merchant, sailing, passenger, medium, tug, and small. There are also 15 fine-grained categories within these basic categories. The VAIS dataset was created by researchers at the University of Washington and the University of California, Berkeley. The dataset was collected from piers in the San Francisco Bay Area. The images in the dataset were captured using two different sensors: a visible-light sensor and an infrared sensor. The visible-light sensor captures images in the visible spectrum, while the infrared sensor captures images in the infrared spectrum. The visible-light sensor is an ISVI IC-C25, which captures 5,056x5,056 bayered color pixel images. The infrared sensor is a Sofradir-EC Atom 1024, which captures 1024x68 pixel images. The VAIS dataset is a valuable resource for object classification research. The dataset is large and diverse, and it includes a variety of ship types and conditions. The dataset is also well-organized and easy to use.

\subsection{Thermal Infrared Video Benchmark for Visual Analysis \citep{TIVBVA}}
The Thermal Infrared Video Benchmark for Visual Analysis (BU-TIV) is a comprehensive dataset designed to facilitate research on object detection, counting, and tracking in single and multiple-view infrared videos. Captured using FLIR SC8000 sensors, the dataset includes over 60,000 frames, hundreds of annotations, and camera calibration files for multi-view geometry. The sequences within the dataset are specifically crafted to test various vision tasks such as tracking single pedestrians or flying bats at low resolution, monitoring multiple moving objects like pedestrians, cars, bicycles, and motorcycles, as well as tracking multiple flying bats and people with the planar motion from multiple views. The dataset also covers 3D tracking of multiple flying bats from three distinct views and counting flying bats in high-density environments. With a diverse range of scenarios and frame sizes, this benchmark dataset is an invaluable resource for researchers and practitioners in the field of visual analysis using thermal infrared videos.

\subsection{Teledyne FLIR Thermal Sensing for ADAS}

The FLIR Thermal Dataset is a free dataset of thermal and visible spectrum images for the development of object detection systems using convolutional neural networks (CNNs). The dataset contains over 26,000 annotated images with 520,000 bounding box annotations captured at day and night, and includes classification of fifteen groups: bike, car, motorcycle, bus, train, truck, traffic light, fire hydrant, street sign, dog, skateboard, stroller scooter, and other vehicle. The frames were captured using a Teledyne FLIR Tau 2 640x512, 13mm f/1.0 (HFOV 45°, VFOV 37°) thermal sensor and a Teledyne FLIR Blackfly S BFS-U3-51S5C (IMX250) camera and a 52.8° HFOV Edmund Optics lens RGB camera. The dataset can be used to train and evaluate object detection algorithms for ADAS and autonomous vehicles.

\subsection{LLVIP A visible-infrared Paired Dataset for Low-light Vision\citep{jia2021llvip}}
The dataset provides visible-infrared paired images for very low-light vision. The dataset has 30976 images, including RGB and IR, translating to 15488 RGB-IR pairs. The wavelength of the dataset is in the thermal range i.e. 8$\sim$14$\mu m$. The dataset is captured using HIKVISION DS-2TD8166BJZFY-75H2F/V2 which is a binocular camera and have both visible light and infrared camera.

\subsection{The TNO Multiband Image Data Collection \citep{tno_dataset}}
The dataset provides intensified visual (390–700 nm), near-infrared (700–1000 nm), and longwave infrared (8–12 µm) nighttime imagery. The dataset contains 16 motion sequences depicting various military and surveillance scenarios, featuring different objects and targets such as people and vehicles set against diverse backgrounds that include both rural and urban settings. The dataset is captured using their own TRICLOBS (TRI-band Color Low-light OBServation) all-day all-weather surveillance system.

\subsection{The Linköping Thermal InfraRed (LTIR) dataset \citep{ltir2015}}
The Linköping Thermal InfraRed (LTIR) dataset is a thermal infrared dataset for evaluation of Short-Term Single-Object (STSO) tracking. It encompasses 20 thermal infrared sequences, each consisting of 536 frames, yielding a substantial volume of data for analysis. With a high-resolution frame size of 1920x480, the dataset offers detailed thermal infrared imagery, enhancing the precision of tracking applications. The dataset was captured using a variety of advanced sensors, including the FLIR A35, FLIR Tau320, and FLIR A655SC. The use of multiple sensors demonstrates the dataset's versatility and suitability for diverse tracking algorithms, making it an invaluable resource for researchers and developers in the field of object tracking using thermal infrared data.

\subsection{ICRA Thermal Infrared Dataset \citep{ICRA_dataset}}
This dataset consists of 4381 aerial thermal infrared images featuring humans, a cat, a horse, and 2418 background images, all of which come with manually annotated ground truths. The image size is 324x256 pixels. The images are split into eight sequences and are available in both 16-bit and downsampled 8-bit formats. The dataset, recorded with a handheld FLIR Tau 320 thermal infrared camera, is suitable for tracking algorithms due to its uniform sampling rate. It also includes a training set comprising cropped images of humans exclusively. Ground truths have been annotated using Matlab evaluation/labeling code (3.2.0).

\subsection{Multispectral Pedestrian Detection Dataset \citep{KAIST}}
The KAIST dataset is a specially curated collection for pedestrian detection tasks, featuring meticulously aligned color-thermal image pairs. The uniqueness of the dataset lies in its use of beam splitter-based specialized hardware, which enables the creation of registered RGB-Thermal images, ensuring a high level of precision and consistency across the dataset. The collection comprises 95,000 images, each having a resolution of 320x256, representing a substantial source of data for analysis. The thermal images in the dataset were captured using the FLIR-A35 sensor, while the RGB images were obtained through the PointGrey Flea3. The blend of these advanced technologies underscores the robustness and reliability of the KAIST dataset in pedestrian detection and related research applications.

\begin{table*}[htbp]
  \centering
  \caption{Tabular representation of Infrared Datasets present in Public Domain}
  \resizebox{\textwidth}{!}{
    \begin{tabular}{|c|p{10.5em}|c|p{6.835em}|p{7.39em}|c|p{16.89em}|}
    \toprule
    \multicolumn{1}{|p{3.72em}|}{\textbf{SNO}} & \textbf{Dataset } & \multicolumn{1}{p{4.555em}|}{\textbf{No of Images }} & \textbf{IR Band } & \textbf{Resolution} & \multicolumn{1}{p{6.165em}|}{\textbf{No of bits }} & \textbf{Sensor} \\
    \midrule
    1     & OSU Thermal Pedestrian Database  & 284   & LWIR  & 360x240  & 8     & Raytheon 300D Thermal Sensor  \\
    2     & OSU Color-Thermal Database  & 17089 & RGB and LWIR  & 320x240  & \multicolumn{1}{p{6.165em}|}{24bit RGB \textbackslash{}\& 8bit thermal } & Sony TRV87 and Raytheon PalmIR 250D thermal sensor  \\
    3     & Terravi Databases  & \multicolumn{1}{p{4.555em}|}{18 Video Sequences } & LWIR  & 320x240  & 8     & L-3 Thermal-Eye 2000AS sensor  \\
    4     & CSIR-CSIO  & \multicolumn{1}{p{4.555em}|}{18 Video Sequences } & LWIR  & 640x480  & \multicolumn{1}{p{6.165em}|}{~ } & ~  \\
    5     & ICRA ASL-TIR  & 4381  & LWIR  & 324x256  & 8     & FLIR Tau 320 thermal infrared camera  \\
    6     & VAIS  & 2865  & RGB and LWIR  & 5056x5056 and 1024x68  & \multicolumn{1}{p{6.165em}|}{~ } & ISVI IC-C25 and  Sofradir-EC Atom 1024  \\
    7     & BU-TIV  & 65590 & LWIR  & upto 1024x  & 16    & FLIR SC8000  \\
    8     & CVC-14  & 8385  & RGB and LWIR  & 476x640  & \multicolumn{1}{p{6.165em}|}{~ } & ~  \\
    9     & CVC-09  & 11071 & LWIR  & 640x480  & \multicolumn{1}{p{6.165em}|}{~ } & ~  \\
    10    & FLIR-ADAS  & 26000 & RGB and LWIR  & 640x512  & \multicolumn{1}{p{6.165em}|}{14bit Tiff, 8bit JPEG , 8bit RGB } & Teledyne FLIR Blackfly S BFS-U3-51S5C (IMX250) camera and Teledyne FLIR Tau 2 thermal sensor  \\
    11    & LLVIP  & 30976 & RGB and LWIR  & 1920x1080 and 1280x720  & \multicolumn{1}{p{6.165em}|}{8 and 16 } & HIKVISION DS-2TD8166BJZFY-75H2F/V2   \\
    12    & TNO   & \multicolumn{1}{p{4.555em}|}{16 Video sequences } & RGB, NIR and LWIR  & 256x256  & 8     & TRICLOBS (TRI-band Color Low-light OBServation) all-day all-weather surveillance system  \\
    13    & LTIR  & 11260 & LWIR  & 1920x480  & \multicolumn{1}{p{6.165em}|}{8 and 16 } & FLIRA35, FLIR Tau320, FLIRA655SC  \\
    14    & KAIST  & 95000 & RGB and LWIR  & 320x256  & 8     & FLIR A35 and PointGrey Flea3  \\
    15    & ATR  & 600000+ & RGB and MWIR  & 640x480  & 16     & Illunis Camera with Nikon Zoom Lens and Night Conqueror MWIR imager   \\
    \bottomrule
    \end{tabular}%
    }
  \label{tab:dataset}%
\end{table*}%

\subsection{Other Datasets}
\subsubsection{MODIS (Moderate Resolution Imaging Spectroradiometer)}
MODIS is a key instrument aboard the Terra and Aqua satellites, providing data in NIR, MWIR, and LWIR regions. It covers a wide range of Earth observation data, including the atmosphere, land, and ocean. 

\subsubsection{Landsat}
The Landsat program is a series of Earth-observing satellite missions jointly managed by NASA and the U.S. Geological Survey. Landsat provides multispectral data, including NIR and SWIR (short-wave infrared) and LWIR bands.

\subsubsection{Sentinel-2}
A part of the Copernicus program, Sentinel-2 is a series of satellites providing high-resolution optical imagery, including NIR and SWIR bands. The data is freely available and frequently used for land monitoring, vegetation, and disaster management applications.

\subsubsection{VIIRS (Visible Infrared Imaging Radiometer Suite)}
VIIRS is an instrument aboard the Suomi NPP and NOAA-20 satellites, providing data in NIR, MWIR, and LWIR regions. It is primarily used for monitoring the Earth's environment, weather, and climate.

\subsubsection{ASTER (Advanced Spaceborne Thermal Emission and Reflection Radiometer)}
A part of NASA's Terra satellite, ASTER provides high-resolution multispectral data, including NIR, SWIR, and TIR (thermal infrared) bands. It is used for studying land surface processes, including vegetation, hydrology, and geology.

\nopagebreak

\section{Synthetic Infrared Scene Simulation Tools}

\begin{table*}[htbp]
  \centering
  \caption{Different elements of Infrared Imaging}
    \begin{tabular}{|c|c|p{13.72em}|}
    \toprule
    \multicolumn{1}{|p{7.72em}|}{\textbf{Target}} & \multicolumn{1}{p{5.835em}|}{\textbf{Atmosphere}} & \textbf{Sensor/Device} \\
    \midrule
    \multicolumn{1}{|p{7.72em}|}{Type} & \multicolumn{1}{p{5.835em}|}{Absorptions} & Spectral Band \\
    \multicolumn{1}{|p{7.72em}|}{Size} & \multicolumn{1}{p{5.835em}|}{Scattering} & Field of View \\
    \multicolumn{1}{|p{7.72em}|}{Emittance} &       & IFOV: Spatial \\
    \multicolumn{1}{|p{7.72em}|}{Reflectance} &       & Resolution System Parameters: MRTD, NETD and MTF \\
    \multicolumn{1}{|p{7.72em}|}{Thermal Contrast} &       & Aperture \\
    \multicolumn{1}{|p{7.72em}|}{Clutter} &       & Detector D* \\
    \multicolumn{1}{|p{7.72em}|}{Motion} &       & Pixel size and pitch \\
    \multicolumn{1}{|p{7.72em}|}{Time} &       & Cold Shield EFF \\
          &       & Signal Processing \\
    \bottomrule
    \end{tabular}%
  \label{tab:addlabel}%
\end{table*}%

The desirable data for developing the Infrared systems is almost impossible to obtain using the measurements. Hence, the Infrared scene simulation is an essential and integral part of any optical system development. Considering its importance, various research has been done to develop GUI-based tools which could simulate any infrared scene simulation with large wavelength variation. These tools generate simulation videos in various steps. 

Most of these tools generate the 3D scene using graphic tools and then use physics-based libraries to model the properties of various elements between the source and the observer/sensor and sensor optics to the detector for complete simulation. The Source-to-Sensor elements include atmospheric properties, spectral irradiance at the source, spectral reflectance at the source, weather conditions, special effects such as fires, smoke, plumes, etc., natural and man-made light sources, dynamic heating and cooling of the source, shadows, etc. The Sensor-to-Detector elements include Optical aberrations, Detector spectral response, Aero-optical effects, detector array, IFOV sampling, Diffractions, Amplification, Gain, Dead Pixels etc. Along with this, they use physics-based material modelling to incorporate the material properties present in the scene. Materials behave differently in different wavelength bands. For example, the signature of a glass is always black in the Infrared region.

Steps involved in the Generation of simulated IR videos
1. Generating the 3D scenes using Graphics tools such as Blender, Unity or Unreal Engine.
2. Finding out the radiance of the object based on the physics-based principles and material properties. The physics-based principles are not limited to Thermophysical properties, Spectral BRDF, Material physical property, spectral signature calculations for various wavelength ranges, Surface temperature properties and their change due to atmospheric conditions, etc. 
3. Atmospheric physical properties modelling using tools such as MODTRAN.
4. Scene rendering based on ray tracing and photon counting algorithms.

There have been many such tools developed, such as OSV \citep{JRMOSV}, Octal \citep{octal}, MuSES and CoTherm \citep{thermoanalytics}, Ondulus \citep{ondulus}, DIRSIG \citep{dirsig}, OSSIM \citep{OSSIM} and others. A brief description of these tools are given below.

\begin{enumerate}
\item JRM Technologies OSV \citep{JRMOSV} is a Real-time spectral EO/IR Sensor Scene Simulator. The OSV can generate real-time scene simulations using in-house developed libraries such as SigSim and SemSim in the 0.2 to 25.0 $\mu m$ wavelength band. The software uses the Open scene graph (OSG) toolkit, which has materially-encoded targets and terrain and SigSim and SenSim libraries to predict correlated radiometrically correct 2D sensor imagery for arbitrary sensor bands under arbitrary weather conditions and spatio-temporal viewing locations. SigSim is a JRM's signature physics library to predict the accurate wavelength band-specific signatures. The SigSim includes the diurnal, ephemeris, scattering parameters, natural and synthetic irradiant causes and surface temperatures to model the wavelength phenomenon accurately. SigSim uses the MODTRAN to model the atmospheric phenomena. SenSim library provides physics-based sensor properties such as post-aperture sensor noise, blur, gain and other effects. Using SenSim, the user can define the optical properties, detector array, signal processing and display parameters.

\item OKTAL-SE's products \citep{octal, LeGoff2000} provides Physics-based rendering engines and 3D geometry-based scene editing for EO/IR applications. They use Blender as their renderer and use advanced thermal equations for IR characterization. They are using MODTRAN for atmospheric modelling. With a dedicated add-on, the tool can generate simulations for complex phenomena such as Sea modelling, Wakes, Foam, cloud layers, 3D clouds, Rain, Snow, Flares, Blasts, Vapor trails, Lights, Exhaust plumes and Lightning strikes. Based on the required scenario, the user can define the Geometry, Physical properties of materials, Atmospheric and thermal conditions, Mobile and instanced objects, Special events during simulation (flares, explosion,…), Trajectories, Scripted animation of the scene and the target using the tool. User can also define the sensor characteristics for sensor-specific simulation and validation.

\item Thermo Analysis's MuSES and CoTherm \citep{thermoanalytics} is a similar EO/IR simulation tool. The two software are used with each other for sensor modelling and scene generation tasks for simulation. The MuSES uses comprehensive heat transfer equations to estimate realistic temperature and EO/IR sensor radiance. It generates high-resolution 3D geometries and calculates component heat sources based on material properties while considering the environmental boundary conditions. The CoTherm provides automation capabilities to the scene simulation process. It helps generate MuSES imagery by taking input parameters from the user.

\item Digital Imaging and Remote Sensing Image Generation (DIRSIG) \citep{dirsig} is a synthetic image generation tool developed by the Digital Imaging and Remote Sensing Laboratory at Rochester Institute. The tool was initially developed for Remote Sensing applications, but later, a flexible development approach was taken to expand its applicability to the development of LIDAR, RADAR, Cloud Modelling, etc. The tool can also be used for low-light-level photon mapping, polarimetric imaging, etc. The tool can produce radiometrically correct broad-band, multi-spectral and hyper-spectral images.

\item OSSIM (Optronic Scene Simulator) \citep{OSSIM, willers2011} is developed by the Council for Scientific and Industrial Research, South Africa and Denel Dynamics Ltd. It is designed explicitly for Infrared Scene simulation and covers the 0.4-20 $\mu m$ of spectral region. The intended use of the OSSIM is for the development of thermal image systems, missile seeker sensors, Sensor and image processing algorithms, etc., along with their optimization and performance simulations. The tool creates radiometrically accurate images for all wavelength bands. They use MODTRAN's Joint Modelling and Simulation Systems (JMASS) interface to model the atmospheric parameters. The tool's thermal equations and heat balance equations are discussed in the next section.

\item CAMEO-SIM for CCD assessment \citep{Moorhead2001Cameosim} was developed by the UK Defence Evaluation and Research Agency and Hunting Engineering Ltd. This physics-based broadband scene simulation tool is specifically designed for evaluating camouflage, concealment, and deception (CCD). It is employed for infrared scene simulation within the 0.4-14 $\mu m$ range. CAMEO-SIM is capable of generating high-fidelity imagery, utilizing an image generator that integrates both radiosity and ray tracing processes. Additionally, it offers the flexibility to produce various levels of fidelity, balancing accuracy and rendering time according to specific requirements.

\end{enumerate}

Due to the ability to generate synthetic data in a wide spectrum band, these tools can be used to generate data for training various deep learning models for applications such as security and surveillance, Target Detection, Tracking, Recognition and Identification, Missile Guidance systems, Automatic Driving Assistance System (ADAS), etc. The deep learning models have changed the course of computer vision with their tremendous efficiency in handling non-linear tasks. These methods learn the data directly and hence require a large amount of data for training and evaluation.  The availability of such data in the Infrared domain is very challenging to find in comparison to RGB data. To develop deep learning models, these tools can be employed for data creation. Further, deep learning is also making these tools better. Since most of the tools rely on graphic tools such as Blender, the scene looks animated and lacks realism. Deep learning methods have lately shown promise in generating realistic images. These proficient models can be utilized in realistic scene creation with these tools. Octal uses such technology to improve its overall simulation experience. 

Most of these tools are proprietary and expensive to purchase. They are designed for vast use cases. They employ a robust verification methodology to ensure the accuracy and integrity of their data and maintain high standards of quality in their processes.

\section{Methods of synthetic IR image/video generation}\label{sec:method}

\begin{figure}
    \centering
    \includegraphics[width=1\linewidth]{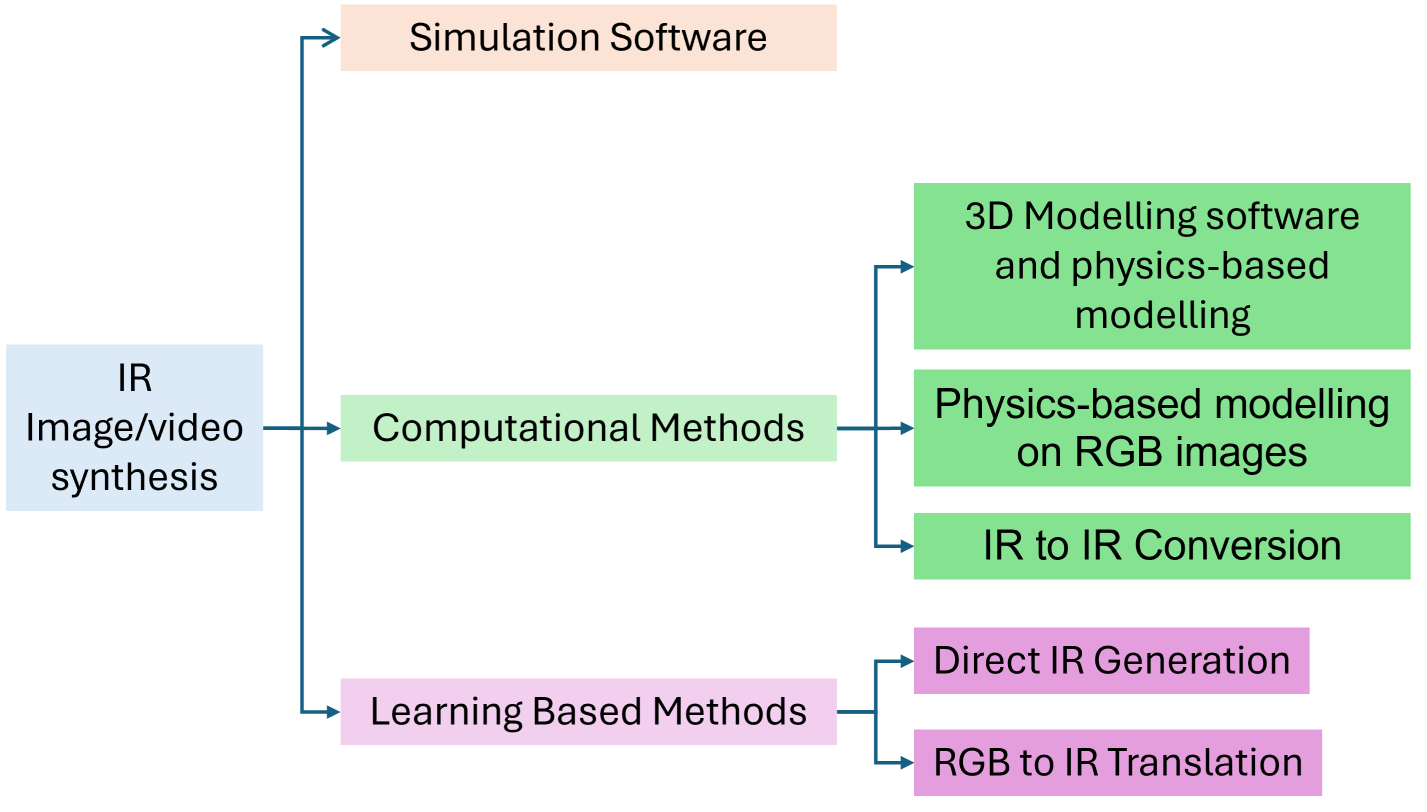}
    \caption{IR image/video synthesis methods classification.}
    \label{fig:classification_blk_diagram}
\end{figure}

In this section, various methods used for the generation of Synthetic IR are discussed. These methods could be categorized into two major parts. First, we are going to discuss the computational methods where IR physics has been prominently used for IR scene simulation, and secondly, we will discuss the latest trends in generative methods in computer vision and their application in synthetic IR scene simulation. A pictorial depiction of the methods discussed in this paper is shown in figure \ref{fig:classification_blk_diagram}.

\subsection{Computational methods}

Computational IR image/video generation methods are among the most reliable as they utilize the physics-based approach to generate realistic and accurate IR imagery. IR signatures of an object depend on multiple factors, including material properties, atmospheric propagation, sensor optics, temperature, reflections, solar influence, viewing angle, shadow effects, and wind speed. The material properties of the object that play a prominent role in its IR signature include emissivity, which is the object's ability to emit thermal radiation. Emissivity is a material property that varies for different materials and alloys, as well as for different wavelength ranges, temperatures, and observation directions. It also depends on the surface structure and regular geometry. The IR photons emitted by the object travel through a medium, the atmosphere. The attenuation caused by the atmosphere changes the behaviour of the IR signature. This attenuation is modeled as atmospheric propagation, which includes ambient temperature, relative humidity, atmospheric temperature, and optical properties of matter in the atmosphere. MODTRAN and related tools are used to model atmospheric propagation. Similarly, once the photon flux reaches the sensor, the sensor optics also modify IR signature. The optics of the sensor, such as lens material, aperture, field of view, transmission, and sensor temperature, play a crucial role in IR imagery. Solar influence is another critical factor that affects the IR signature. The sun's rays falling on the object carry heat, which is absorbed, reflected, and radiated by the object. These rays are also refracted by the atmosphere and fall on the detector, significantly changing the IR imagery. The shadow effect, which blocks the fall of solar rays on the object, changes the heat equilibrium of that part and thus changes its IR signature. 

For radiometrically accurate computation method based IR scene generation, researchers have explored multiple methods. A seminal paper by Garnier et al. \citep{Garnier1999Infrared} introduced a sensor modeling methodology that establishes the geometric and radiometric relationships between points in a 3D observed scene and the corresponding pixels in the IR sensor output image. This work provided the mathematical foundation for physics-based sensor modeling techniques. \citep{Jiang2019, Zhijian2022, Dulski2011, Qi2019, LI2015533, Jiang2003dynamicir} have used 3D modelling software to create a 3D Scene and then using physics-based modelling, they have generated the IR scenes. \citep{Kim2014, Bae2018, Bae2019} have used a known IR signature from a specific band and using the temperature calculation method they have produced the IR signature in different spectral bands. \citep{shankarmore, Leja2022, Choi2009, Yu1998} have used images and developed physics-based modelling to describe the IR signature emitted by the objects in the image.

\begin{table*}[htbp]
  \centering
  \caption{List of Computational Methods}
    \begin{tabular}{|c|c|}
    \toprule
    \textbf{Method} & \textbf{Papers} \\
    \midrule
    \multicolumn{1}{|c|}{\multirow{4}[8]{*}{3D Modelling software and physics-based modelling}} &  Zhaoyi et al. \citep{Jiang2003dynamicir}\\
    \cmidrule{2-2}          & Jiang et al.\citep{Jiang2019} \\
\cmidrule{2-2}          & Zhijan et al. \citep{Zhijian2022} \\
\cmidrule{2-2}          & Dulski et al. \citep{Dulski2011}\\
\cmidrule{2-2}          & Qi et al. \citep{Qi2019}\\
\cmidrule{2-2}          & Li et al. \citep{LI2015533}\\
    \midrule
    \multicolumn{1}{|c|}{\multirow{3}[6]{*}{IR to IR Conversion}} & Kim et al. \citep{Kim2014}\\
\cmidrule{2-2}          & Bae et al. \citep{Bae2018}\\
\cmidrule{2-2}          & Bae et al. \citep{Bae2019}\\
    \midrule
    \multicolumn{1}{|c|}{\multirow{4}[8]{*}{Physics-based modelling on RGB images}} & Choi et al. \citep{Choi2009}\\
\cmidrule{2-2}          & Yu et al. \citep{Yu1998}\\
\cmidrule{2-2}          & More et al. \citep{shankarmore}\\
\cmidrule{2-2}          & Leja et al. \citep{Leja2022}\\
\cmidrule{2-2}          &  Zhaoyi et al. \citep{Jiang2003dynamicir}\\
    \bottomrule
    \end{tabular}%
  \label{tab:dd}%
\end{table*}%

Yu et al. \citep{Yu1998} have proposed a method for realistic IR image generation by establishing a heat equilibrium equation of the object surface and using it along with the physics of the heat transfer inside and on the boundary of the object; they computed the temperature and radiometric details. They applied the method in a patch-wise manner on the object and used Gouraud shading along with corresponding radiometric information to draw each patch in IR. They have calculated the radiance of the object's surface in the atmospheric window where the atmosphere's attenuation is minimal to certain specific wavelengths. They have not used the MODTRAN for atmospheric attenuation. Even though the paper thoroughly explores the physics behind various heats occurring due to solar irradiance and the object itself, their IR signature is dependent on the assumption of the transparent atmospheric window, which is not practical in real life.

Choi et al. \citep{Choi2009} proposed composite heat transfer model for objects present in the scene to calculate surface temperature distribution. The heat transfer model included the conduction, convection and solar irradiance for calculation of surface temperature. The  equation governing the surface temperature is similar to equation \ref{heat_equilibrium_temperature}. They have used MODTRAN 4 for calculating the atmospheric transmittance, various radiance such as atmospheric background radiance, solar and lunar radiance, thermal radiance, etc. To calculate the radiance received by the sensor, they have considered the emission from the object's surface, solar irradiance reflection by the object surface, and atmospheric scattering. The resulting equation is similar to the big equation discussed in sections before. They have demonstrated the efficiency of their computational model by generating the infrared signature of objects made up of Asphalt and Aluminium. They have calculated the diurnal Infrared signature for these objects. The Density, specific heat, thermal conductivity, and total absorptivity were essential material properties used during radiance calculations. 

Leja et al. \citep{Leja2022} proposed a mathematical model and IR data generation for uncooled IR sensors with the aim of aiding the calibration of IR sensors. IR sensors are prone to fixed pattern noise (FPN) caused by the non-uniform current-voltage characteristics of the amplifier and bolometer's non-uniform responsiveness. This results in undesirable artefacts in the imagery, including dead and defective pixels. They have used the mathematical modelling of various elements of sensors, including bolometer, focal plane array, optics and environment, to generate synthetic infrared images pertaining to an uncooled sensor. With mathematical modelling, they were able to simulate various sensor issues such as pixel defectiveness, non-uniformity, etc.

More et al. \citep{shankarmore} have proposed synthetic IR cloud generation, radiance calculation of aircraft and scene rendering for air-borne targets. They have used MODTRAN for radiometric calculation and Virtual reality modeling language for scene rendering. They proposed a method to generate clouds with rich spectral information and texture by improving on the Gardner's method and self-similarity algorithms. They have used 3D geometric models of aircraft with planar triangular facets and assumed the temperature of various elements of the aircraft and then used MODTRAN to calculate the radiance of the aircraft as a whole.

\begin{figure*}
    \centering
    \includegraphics[width=1\linewidth]{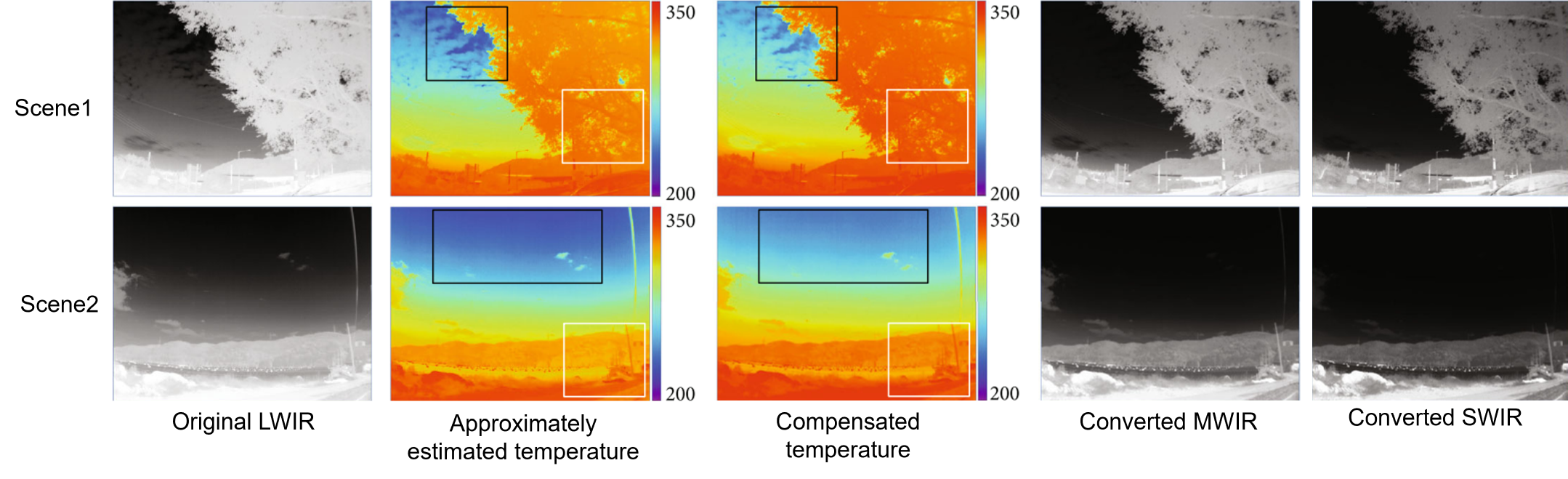}
    \caption{Conversion of one IR band to another IR band by first estimating the temperature of objects in the image, then compensating the temperature and predicting IR signature in different bands using method proposed by Bae et al \citep{Bae2019}. Image courtesy \citep{Bae2019}}
    \label{fig:bae2019}
\end{figure*}

Another approach taken by researchers is to generate IR images of different wavelength bands from an IR image of a known wavelength band. These methods can also be called conversion methods, which convert IR images from one wavelength band to another. Kim et al. \citep{Kim2014} and Bae et al. \citep{Bae2018, Bae2019} have proposed the generation of IR images of arbitrary spectral bands using IR images with known spectral bands. They first estimate the radiance of the target and background from the IR images of arbitrary wavelength bands. Using the radiance, they estimate the temperature component of objects in the IR images. Using this information, they have generated the temperature image. Further, using the temperature-to-radiance models and radiance to grey-level transfer functions, they generated the image in three different bands, namely LWIR, MWIR and SWIR. The result obtained by the method is shown in figure \ref{fig:bae2019}.

\begin{figure}
    \centering
    \includegraphics[width=1\linewidth]{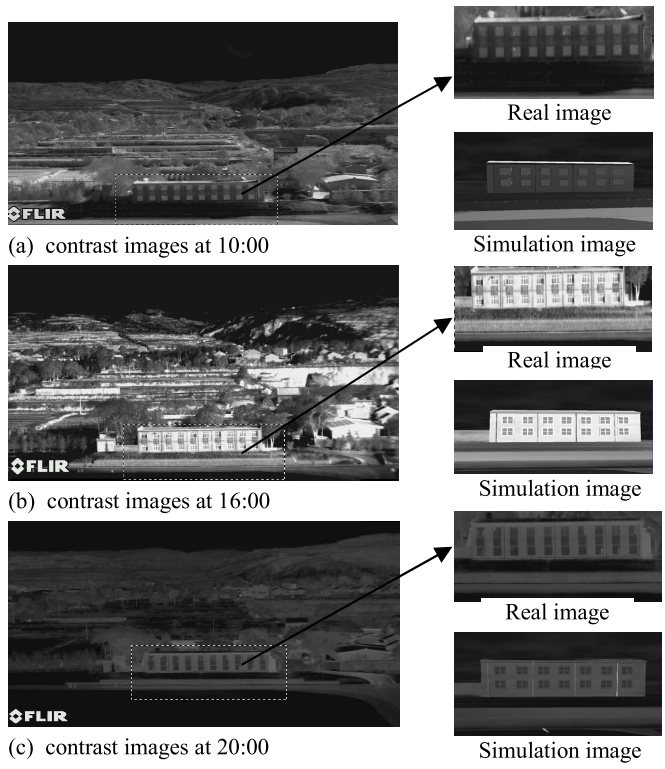}
    \caption{IR scene simulation results using OGRE rendering engine and physics-based model proposed by \citep{He2013}. Image courtesy \citep{He2013}.}
    \label{fig:he2013_results}
\end{figure}

Computer graphic-aided tools are also used by the researchers for the generation of accurate synthetic scenes. These methods serve as the foundation for the development of software-based IR generation methods. He et al. \citep{He2013} have used the Object-oriented Graphics Rendering Engine (OGRE), an open-source engine, to generate the 3d geometrical models. Further, they have calculated the infrared radiation and texture of the models using thermal models. Atmospheric effects and radiation transfer model effects were then added to these textures to accommodate atmospheric transmission of the radiation in a specific wavelength. Then the 3D infrared scene was rendered using the final textures of the 3D objects. Results of the method is shown in figure \ref{fig:he2013_results}. Similar to this, Jiang et al. \citep{Jiang2019} have used heat balance equation models and OSG engine to predict the surface temperature of various materials and radiance and generate large-scale IR scenes, respectively. The IR image was generated using the ray tracing method. Further, they have used the material partitioning method to improve the overall temperature prediction.

\begin{figure}
    \centering
    \includegraphics[width=1\linewidth]{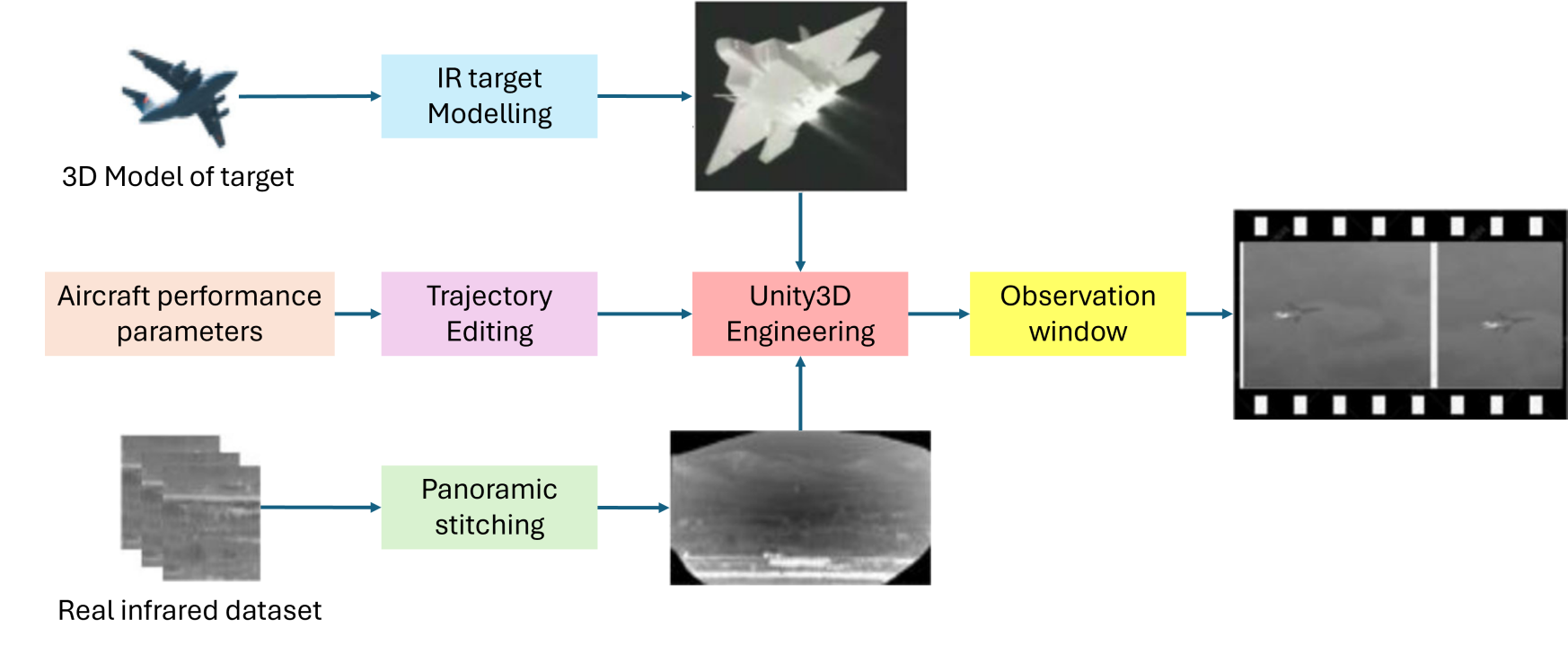}
    \caption{Zhijian et al. \citep{Zhijian2022} pipeline for generating synthetic IR scenes with embedded targets: This pipeline leverages physics-based modelling to accurately simulate target thermal signatures and utilize the Unity rendering engine to define target trajectories and other scene elements. The resulting output is a collection of synthetic IR scenes with seamlessly integrated targets.}
    \label{fig:zhijian_2022}
\end{figure}

\begin{figure}
    \centering
    \includegraphics[width=1\linewidth]{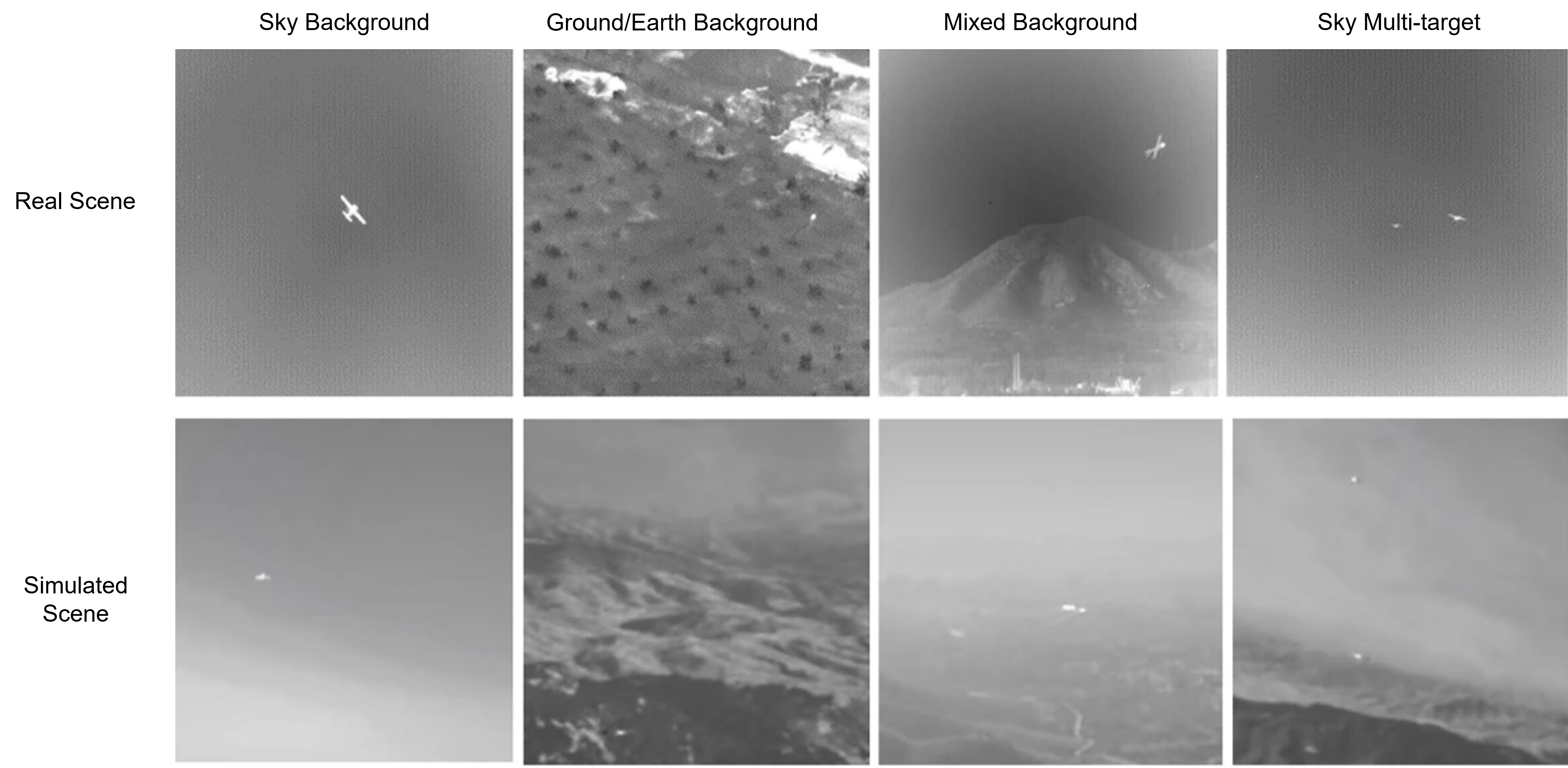}
    \caption{Real and simulated Infrared Scenes with targets obtained using the graphic rendering tool and physical simulation. Image courtesy \citep{Zhijian2022}}
    \label{fig:Zhijian_2022_result}
\end{figure}

Zhijian et al. \citep{Zhijian2022} have used Unity3D graphic rendering engine to generate the IR imagery for simulation. They have used panoramic IR images generated using stitching small real IR images as a background. They mathematically modelled the 3D aircraft trajectory and altitude. Further, they have used IR physical modelling to generate the target. The 3D aircraft's tail nozzle, tail flame and skin have been modelled as the target. Then, all this information is fused using the Unity3D graphic rendering engine to generate Infrared data with target tracking. The proposed pipeline is shown in figure \ref{fig:zhijian_2022} and the resultant simulated scenes are depicted in figure \ref{fig:Zhijian_2022_result}.

Dulski et al. \citep{Dulski2011} have proposed a numerical modelling method for simulating cloud radiation in the infrared spectral range for development of automatic target recognition algorithms. The method incorporates experimental data based on characteristic temperature profiles of clouds and sky, considering different seasons and meteorological conditions in Poland. The researchers developed the IR Sky software to generate virtual thermal images of sky and clouds, which were then analyzed to understand the impact of radiation features on the thermo-detection process. 

Computer graphics-based IR generation systems provide accurate and efficient ways to model realistic scenes but are not specifically designed to handle radiative transfer simulations. Qi et al. \citep{Qi2019} have proposed a photon tracing method and backward path tracing method to simulate BRDF and photon flux information for multiple spectra and to generate sensor and large scale spectral images respectively. The backward path tracing is also used to simulate thermal infrared radiation by computing the sunlit and shaded scene components. 


\subsection{Learning based methods}

\begin{table*}[htbp]
  \centering
  \caption{List of Learning based Methods}
    \begin{tabular}{|l|l|l|}
    \toprule
    \multicolumn{1}{|c|}{\textbf{Method}} & \multicolumn{1}{c|}{\textbf{Papers}} & \multicolumn{1}{c|}{\textbf{Deep Learning Technique}} \\
    \midrule
    \multicolumn{1}{|c|}{\multirow{15}[30]{*}{RGB-IR conversion}} & Kniaz et al.\citep{Kniaz2017THERMALNETAD} & SqweezeNet CNN \\
\cmidrule{2-3}          & Kniaz et al.\citep{Kniaz2019} & ThermalGAN \\
\cmidrule{2-3}          & Zhang et al.\citep{Zhang2019} & CycleGAN and Pixel GAN \\
\cmidrule{2-3}          & Mizginov et al.\citep{Mizginov2019} & Vanila GAN with Unet as Generator \\
\cmidrule{2-3}          & Li et al.\citep{Li2019} & Multi-Generator Network \\
\cmidrule{2-3}          & Qian et al.\citep{Qian2020} & Pix2Pix GAN with Sparse Unet as Generator \\
\cmidrule{2-3}          & Abbott et al.\citep{Abbott2020unsupervised} & CycleGAN with object specific loss \\
\cmidrule{2-3}          & Yuan et al.\citep{Yuan2020} & Conditional GAN\citep{mirza2014conditionalgenerativeadversarialnets} with Robust adaptive loss function \citep{mirza2014conditionalgenerativeadversarialnets} \\
\cmidrule{2-3}          & Uddin et al.\citep{Uddin2021} & Attention GAN \\
\cmidrule{2-3}          & Li et al.\citep{Li2021} & Pix2Pix GAN with D-link Net as Generator \\
\cmidrule{2-3}          & Wang et al.\citep{Wang2022} & CPSTN-style transfer Network \\
\cmidrule{2-3}          & Kim et al.\citep{Kim2022} & BiCycleGAN (Intensity Modulation Network) \\
\cmidrule{2-3}          & Ulusoy et al.\citep{ULUSOY2022} & Vanila GAN with Unet as Generator \\
\cmidrule{2-3}          & Kim et al.\citep{Kim2022_2} & CNN based neural style transfer network \\
\cmidrule{2-3}          & \"{O}zkano{\u{g}}lu et al.\citep{zkanolu2022} & InfraGAN \\

\cmidrule{2-3}          & Li et al.\citep{Li2023DAGAN} & DAGAN \\

\cmidrule{2-3}          & Yi et al.\citep{Yi2023Cycle} & CycleGan architecture with channel and spatial attention \citep{woo2018cbam} and \\ 
 & & gradient normalization module \citep{wu2021gradient} \\
\cmidrule{2-3}          & Uddin et al.\citep{Uddin2023} & MWIR-GAN \\
\cmidrule{2-3}          & Wang et al.\citep{Wang2023} & V2IR-GAN \\
    \midrule
    Direct IR Generation & Zhang et al.\citep{Zhang2019_2} & SIR-GAN \\
    \bottomrule
    \end{tabular}%
  \label{tab:addlabel}%
\end{table*}%

Deep learning based image generation methods recently have shown very promising result in RGB based image generation. These methods are fast as compared to the computational methods and produce more diversified images. There have been multiple methods to generate RGB images, including Deep Boltzmann Machines, which are energy-based models; Variational Autoencoders \citep{Wei2020, kingma2013}, which are directed probabilistic graphic models; Deep AutoRegressive models; normalising flow Models and Generative models \citep{goodfellow2014, mirza2014conditionalgenerativeadversarialnets, Zhu2017, Isola2017}. 

Deep learning-based IR generation models can be divided into two groups. The first group consists of models which work on translating the RGB images to IR images \citep{Kniaz2017THERMALNETAD, Kniaz2019, Zhang2019, Mizginov2019, Li2019, Qian2020, Uddin2021, Li2021, Wang2022, Kim2022, ULUSOY2022, Kim2022_2, zkanolu2022, Uddin2023, Wang2023, Yuan2020, Yi2023Cycle, Li2023DAGAN, Abbott2020unsupervised}. These models require RGB-IR pair data for training and work on the principle of mapping RGB images to IR. The second group consists of direct IR generation \citep{Zhang2019_2}. These models are trained on IR images directly and generate similar IR images from the learned distributions.

Generating thermal images from the RGB image is particularly an inverse and ill-posed task. Since the RGB image contains radiative information in the range of 400-700 nm, whereas the thermal images belong to a band of 3-15 \(\mu\)m, most of the models have tried this as a problem of domain translation instead of developing physics-based models. 


Mizginov et al. \citep{Mizginov2019} have explored the multimodal network for synthetic IR generation from color band to thermal band. Researchers argue that GAN networks average the thermal contrast over the entire object, hence losing the important thermal characteristics of particular regions. To address this issue, they provide segmented thermal zones of objects along with the depth map to improve the network's ability to localise the object's thermal characteristics in the generated image, thereby improving the overall synthetic IR quality. They have used 3D modelling tools to create realistic 3D models and generated thermal contrast maps, depth maps and masks of objects. These were given as input to the GAN network with U-net \citep{unet} as the generator and patchGAN \citep{patchgan} as the discriminator. \"{O}zkano{\u{g}}lu et al. \citep{zkanolu2022} have proposed an InfraGAN network which uses a Unet-based generator to learn the transfer mapping between the IR and RGB. The generator loss is optimized using structural similarity losses (SSIM) and L1 loss, and discriminator loss. The discriminator proposed in the method classifies the entire image as real or fake and then classifies each pixel as real or fake. Kniaz et al. \citep{Kniaz2017THERMALNETAD} have used the squeezeNet CNN network to translate the RGB data into thermal images. The network was inspired by colorization problem. They developed a training dataset consisting of 1000 images of geometrically aligned RGB and thermal images of various objects. Yuan et al. \citep{Yuan2020} have used conditional GAN \citep{mirza2014conditionalgenerativeadversarialnets} for synthesizing the NIR images from RGB images. They have used a robust adaptive loss function proposed in \citep{barron2019generaladaptiverobustloss} along with the SSIM loss. They have used paired images from the Sentinel-2 dataset.

\begin{figure}
    \centering
    \includegraphics[width=1\linewidth]{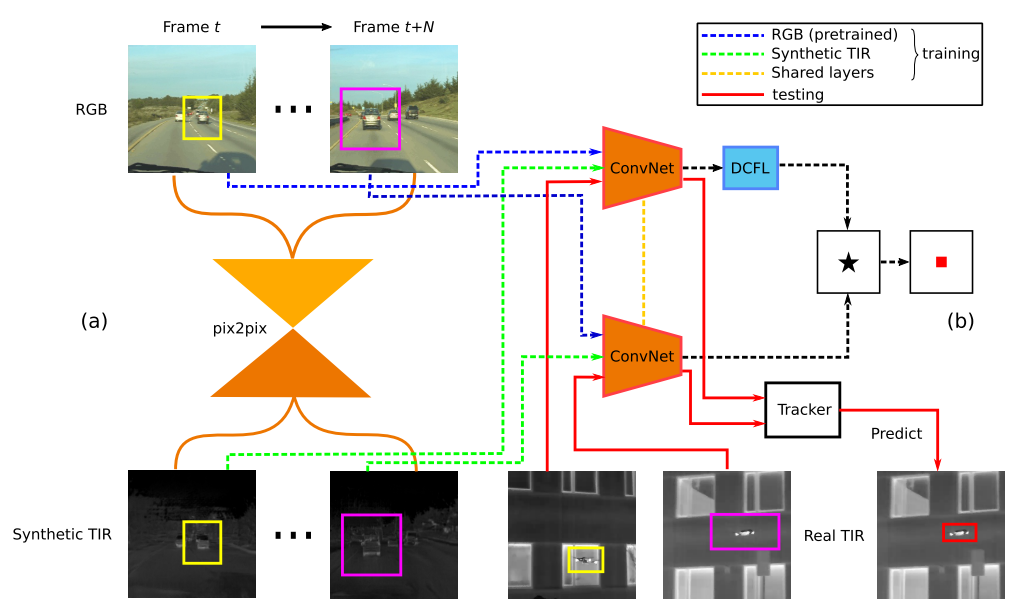}
    \caption{Pix2Pix Architecture presented by Zhang et al \citep{Zhang2019}. The (a) section of the image represents the image-to-image translation component which was used for generating synthetic TIR tracking dataset. The (b) section in the image represents the tracking pipeline which uses the feature from the (a) block for effective tracking. Image courtesy \citep{Zhang2019}.}
    \label{fig:zhang_arch2019}
\end{figure}

\begin{figure}
    \centering
    \includegraphics[width=1\linewidth]{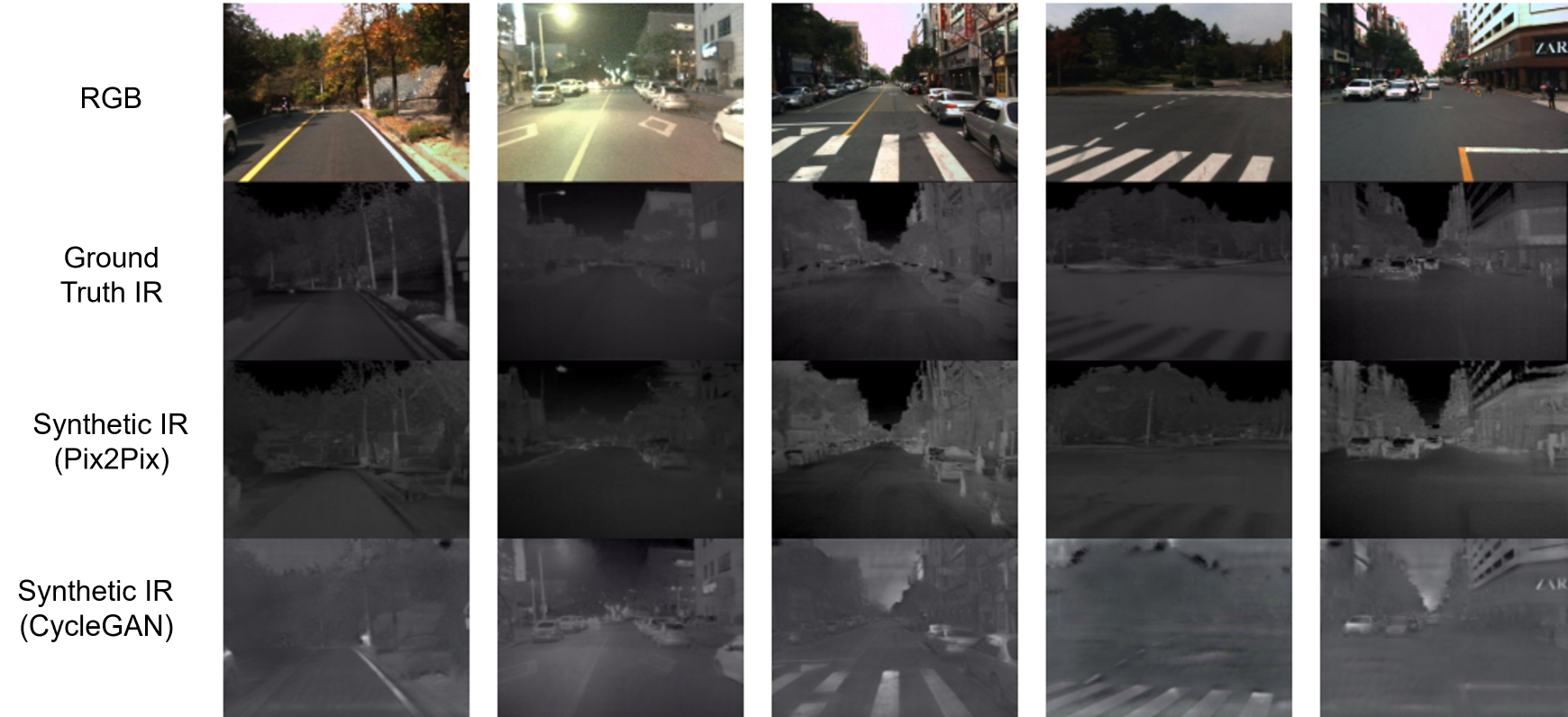}
    \caption{Pix2Pix and CycleGAN based image-to-image translation task on KAIST\citep{KAIST} dataset, as performed by \citep{Zhang2019}. Image courtesy \citep{Zhang2019}.}
    \label{fig:zhang_result2019}
\end{figure}

Pix2Pix and CycleGAN based network has been the foremost choice of researchers due to their direct application in domain adaptation and image translation. Zhang et al. \citep{Zhang2019}, shown in figure \ref{fig:zhang_arch2019}, explored image-to-image translation models such as CycleGAN\citep{Zhu2017} and Pix2Pix GAN\citep{Isola2017} for unpaired and paired image translation to create large IR tracking data from RGB videos. Further, the intermediate features of these networks were used to enhance the tracking of IR images. The results obtained by these models as presented in their paper is shown in the figure \ref{fig:zhang_result2019}. Similar to these architecture Qian et al. \citep{Qian2020} have proposed sparse U-net generator\citep{unet} and patchGAN based Pix2Pix network for RGB to Thermal IR images. They select partial low-level and high-level features only and use intensity and gradient loss to optimise the network. Li et al. \citep{Li2021} have used D-LinkNet architecture instead of U-net as their generator in the Pix2Pix network to learn the image textures and interdependencies within the image to improve overall generated synthetic IR quality.

\begin{figure}
    \centering
    \includegraphics[width=1\linewidth]{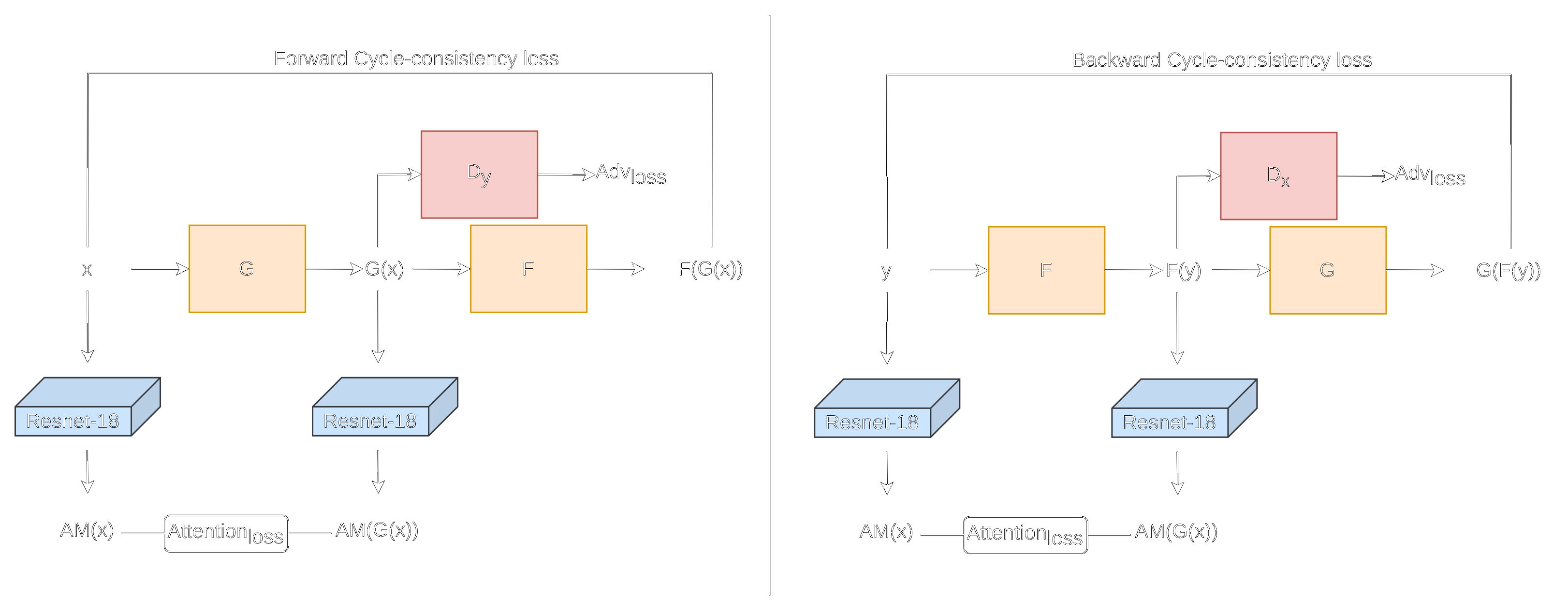}
    \caption{Attention Gan architecture \citep{Uddin2021}.}
    \label{fig:attnGAN}
\end{figure}

\begin{figure}
    \centering
    \includegraphics[width=1\linewidth]{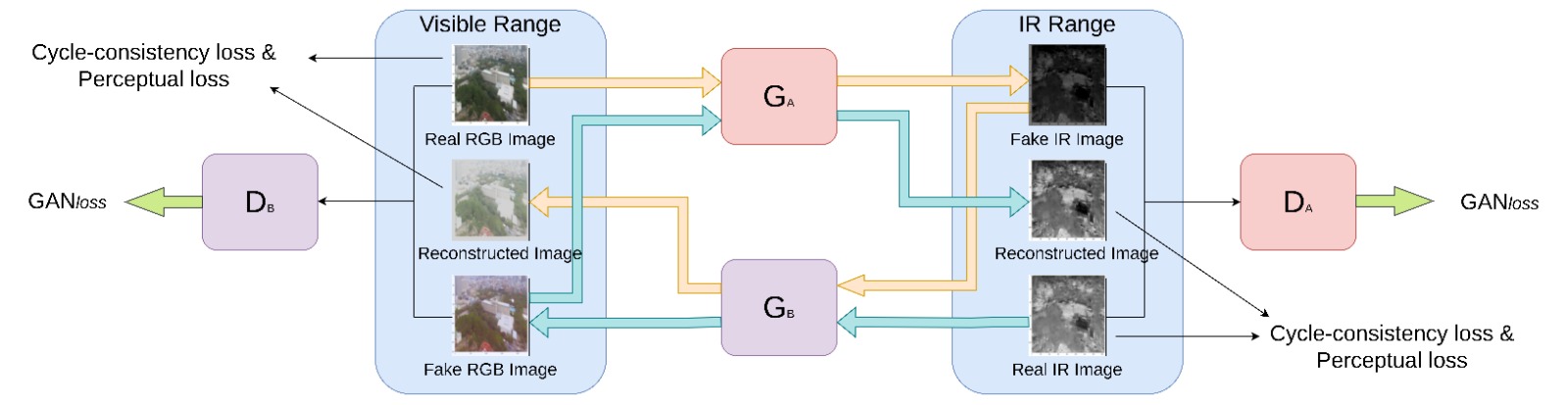}
    \caption{MWIRGAN architecture \citep{Uddin2023}}.
    \label{fig:MWIRGAN}
\end{figure}

Abbott et al. \citep{Abbott2020unsupervised} have used the CyclgeGAN \citep{Zhu2017} network along with a newly proposed loss function which is object specific loss and calculated as absolute distance between the region of interest detected in fake and real RGB and IR images using the detector. This loss is designed based on the features of night vision IR images and RGB vision images. Uddin et al. \citep{Uddin2021} used attention GAN, shown in figure \ref{fig:attnGAN}, to focus on the target areas, preserving the target areas' correctness. Their network is similar to the CycleGAN, except they used two ResNet18-based teacher networks, which generate the attention maps and are used to teach both discriminators and generator networks to focus on the target area. The authors have demonstrated significant improvement in generated synthetic IR images over the CycleGAN \citep{Zhu2017} and CatGAN networks. The teacher network is pre-trained to classify a variety of military vehicles, making it learn to generate attention maps of vehicles specifically. Then, this network is used in the main architecture. They have used the DSIAC dataset \citep{ATR} for training their network. Yi et al. \citep{Yi2023Cycle} have argued that most of the domain adaptation methods are incapable of perceiving the salient regions in the RGB images and hence cannot effectively generate textures details in the synthesized IR images. To address this issue, they have proposed a gradient normalization technique based CycleGAN \citep{Zhu2017} architecture to generate texture rich IR images while effectively stabilizing the GAN during training. They have used channel and spatial attention as proposed in \citep{woo2018cbam} to effectively model the images. Li et al. \citep{Li2023DAGAN} have proposed a Dual attention GAN (DAGAN) Network, to translate RGB images to thermal domain. They have used DAGAN to segregate foreground and background features and learn attention on these to improve the quality of translation to thermal images. They have further validated the data using different fire tests with different setups. They have also used cyclic loss proposed in \citep{Zhu2017}.

\begin{figure}
    \centering
    \includegraphics[width=1\linewidth]{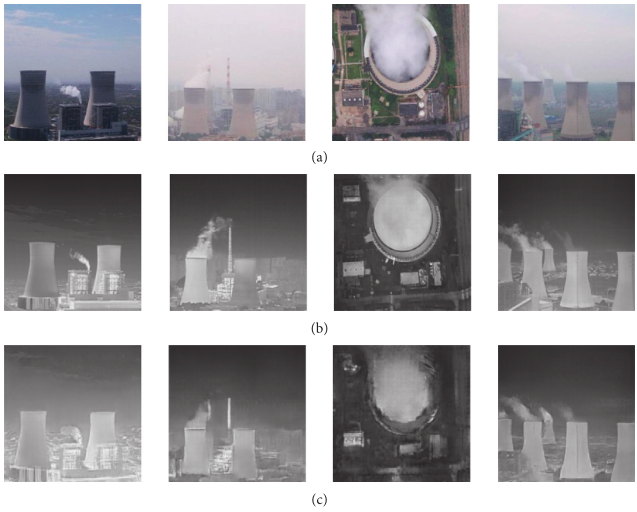}
    \caption{Visual Results of RGB to IR translation from IGAN \citep{Li2021}. (a) represents the RGB image, (b) is Real IR and (c) is the generated IR of the cooling tower from the IGAN model. Image courtesy \citep{Li2021}.}
    \label{fig:enter-label}
\end{figure}

The authors in \citep{Kniaz2019} have used GAN-based architecture, ThermalGAN, to convert RGB images to LWIR images for person re-identification. They have used a derivative of the BiCycleGAN \citep{bicyclegan} framework where the first GAN predicts the thermal segmentation maps from the input RGB image, and the second GAN predicts the relative local temperature contrasts. Then, these two were combined to get a final thermal image. They have also proposed a Thermal world dataset for the person reidentification task in the paper. Kim et al. \citep{Kim2022} have proposed a twofold approach to generate synthetic IR with a target for target tracking applications. They have used BicycleGAN \citep{bicyclegan} based network for IR synthesis from RGB image calling it as background images and then used intensity modulation network along with the target masked background images to render realistic IR targets on them. The proposed intensity modulation network is a variant of GAN that is used to adapt the target mask region to the background region. They have trained the BiCycleGAN on the KAIST dataset \citep{KAIST}, which features aligned color-thermal image pairs.

Wang et al. \citep{Wang2022} have proposed a cross-modality perceptual style transfer network to generate pseudo IR images from RGB images which have sharp structure of the actual IR image. Building on the idea that the pseudo IR and RGB images will be paired, they have used these pseudo IR images to calculate the displacement between the RGB image and Real IR images for registration and fusion purposes. The perceptual style transfer constraint controls the learning of the network to generate structure enhanced pseudo Infrared images. 

IR scenes have unique distinctive characteristics based on many factors. This complicates the learning process as a single generator could not model all unrelated characteristics of the IR scene. Learning all these characteristics from the RGB image to generate a realistic IR image is very difficult. To address this issue, Li et al. \citep{Li2019} have proposed an ensemble learning-based multi-generator network learning different semantic information for IR synthesis for remote sensing applications. They have used a ResNet-50-based scene classification prior to the multi-generator network to make sure that each generator learns the generation of the scene with a specific characteristic. To avoid the complexity of the network arising due to a multi-generator network they have used cluster-based methods to identify and group the characteristics which are similar in nature and can be addressed by a single generator hence significantly limiting the number of generator required. They have used Pix2Pix GAN network for image translation.

Most of the Deep learning based IR generation models do not explore the physical principles while generating the IR images. To address this issue, Wang et al. \citep{Wang2023} have proposed V2IR-GAN to generate IR images which models the physical process of IR images while generating novel images.  They have developed three modules: the spontaneous emission module, the reflected radiation module and the transmission coefficient module, which models the spontaneous emission generated by the object, inter-reflection radiation from the surrounding object or environment, and atmospheric radiations, respectively. They have developed a more physics-based approach to generate the thermal images from the RGB images.


Zhang et al. \citep{Zhang2019_2} have proposed IR image refinement network SIR-GAN which learns the bidirectional mappings between two domains Real IR and computationally calculated synthetic IR for enhancing the realism in the simulated IR images. They have used cycle consistency loss, SIR refinement loss and adversarial loss to optimize there network. They have used FlexCam expert IR thermal imager to capture 800 samples of real IR images. To generate synthetic IR images, they have first build the 3D geometric model of the target and then used IR physical modeling to get IR textures of the target. Further, they have used OGRE rendering along with the atmopsheric model to simulate the IR images. Once these synthetic images were obtained they used SIR-GAN to bring realism into the synthetic IR images.

The constraints of having a sufficient supply of accurately labelled images can often hinder the training of deep learning systems, leading to insufficient diversity in aspects like target angles, time of the day, and seasonal variations. This challenge becomes particularly pronounced when working beyond the visible spectrum. Acquiring an abundant collection of real remote-sensing images can be costly, demanding extensive fieldwork and post-processing efforts. Thankfully, synthetic imagery emerges as a valuable and cost-effective substitute.

\nopagebreak

\section{Challenges in Synthetic Infrared Image synthesis}
\label{sec:challenges}

Despite the remarkable potential of Synthetic Infrared (IR) Image synthesis, it's important to note that the technology is not without its challenges. These hurdles often revolve around the accuracy and realism of the synthetic images, the complexity of the modeling process, and ethical considerations.

\textbf{Accuracy and Realism:} One of the main challenges in synthetic IR image synthesis is ensuring the images accurately represent real-world thermal scenarios. This involves not only replicating the appearance of different objects under IR imaging but also simulating the various factors that can affect an object's thermal signature, such as its material properties, the environmental conditions, or the angle of observation.

\textbf{Data Diversity:} To train robust and reliable models, a wide range of scenarios need to be represented in the synthetic IR images. This includes different weather conditions, times of day, seasons, and types of objects. Ensuring such diversity in the synthetic dataset can be a challenging task.

\textbf{Computational Complexity:} The process of generating synthetic IR images can be computationally intensive, requiring significant processing power and time. This can be a barrier, especially when large datasets of synthetic images are needed for training robust machine learning models.

\textbf{Validation:} Validating the effectiveness of synthetic IR images can be a complex process. This usually involves comparing the performance of models trained on synthetic images with those trained on real-world IR images. However, obtaining a large and diverse dataset of real-world IR images for this comparison can be difficult due to privacy, security, and logistical issues.

\textbf{Ethical Considerations:} There are also ethical considerations involved in the use of synthetic IR images. For instance, if used in surveillance or facial recognition systems, there are concerns about privacy and consent. Additionally, there is the potential risk of misuse if the technology falls into the wrong hands.

\textbf{Generalizability:} Models trained on synthetic IR images need to generalize well to real-world situations. However, due to the inherent differences between synthetic and real-world images, there can be a domain gap that hampers the model's performance in real-world scenarios.

\section{Conclusion}
\label{sec:conclu}

Synthetic infrared image and video generation is a crucial research area with significant implications across myriad fields, including defense, gaming, healthcare, and environmental monitoring. This comprehensive survey offers an in-depth overview of the existing methods for creating synthetic IR imagery, encompassing the physics of IR radiation emission, sensor modeling, atmospheric attenuation, simulation tools, and computational techniques for IR scene simulation. Additionally, it addresses various datasets available in the infrared domain used for IR image generation through deep learning methods, as well as the deep learning-based approaches for IR scene creation.

The survey identifies the challenges in IR scene generation, such as the need for more realistic and diverse datasets, the demand for more accurate and efficient simulation tools, and the necessity of addressing the domain shift problem between synthetic and real-world IR data. Moreover, the development of synthetic IR imagery has the potential to enable new applications in autonomous systems, surveillance, and predictive maintenance, significantly impacting various industries.

Future research directions may include the advancement of more sophisticated simulation tools, the creation of larger and more varied datasets, and the exploration of innovative deep learning architectures and techniques for IR scene generation. Additionally, integrating synthetic IR imagery with other modalities, such as visible and hyperspectral imaging, can lead to more robust and accurate scene understanding and object recognition. Overall, this survey aims to provide a foundation for further research and development in the field of synthetic IR image and video generation, which has the potential to transform numerous industries and applications.













\printcredits

\bibliographystyle{cas-model2-names}

\bibliography{main}



\end{document}